\documentclass{article}
\usepackage{ijcai26}

\usepackage{times}
\usepackage{soul}
\usepackage{url}
\usepackage[hidelinks]{hyperref}
\usepackage[utf8]{inputenc}
\usepackage[small]{caption}
\usepackage{graphicx}
\usepackage{amsmath}
\usepackage{amsthm}
\usepackage{booktabs}
\usepackage{algorithm}
\usepackage{algpseudocode}
\usepackage[switch]{lineno}
\usepackage{enumitem}
\usepackage{amssymb}
\usepackage{pifont}

\usepackage{pgfplots}
\pgfplotsset{compat=1.18} % Or 1.18, 1.9 depending on your Overleaf version

\definecolor{oursred}{RGB}{214, 39, 40}
\definecolor{gresblue}{RGB}{31, 119, 180}
\definecolor{lisatorange}{RGB}{255, 127, 14}
\definecolor{gsvagreen}{RGB}{44, 160, 44}
\definecolor{textviolet}{RGB}{148, 103, 189}

\usepackage{tikz}
\usetikzlibrary{calc} % REQUIRED for the coordinate math ($ ... !0.5! ... $)
\usetikzlibrary{patterns}
\usepackage{multirow}
\usepackage{graphicx}

\usepackage{amsmath,amsfonts,bm}

\def\eqref#1{equation~\ref{#1}}
\def\1{\bm{1}}

\newcommand{\cmark}{\ding{51}}
\newcommand{\xmark}{\ding{55}}

\def\rvp{{\mathbf{p}}}

\def\rvr{{\mathbf{r}}}
\def\rvs{{\mathbf{s}}}
\def\rvt{{\mathbf{t}}}

\def\rvx{{\mathbf{x}}}
\def\rvy{{\mathbf{y}}}

\DeclareMathAlphabet{\mathsfit}{\encodingdefault}{\sfdefault}{m}{sl}
\SetMathAlphabet{\mathsfit}{bold}{\encodingdefault}{\sfdefault}{bx}{n}

\def\gD{{\mathcal{D}}}

\def\gL{{\mathcal{L}}}

\newcommand{\R}{\mathbb{R}}

\title{MAgSeg: Segmentation of Agricultural Landscapes in High-Resolution Satellite Imagery using Multimodal Large Language Models}

\author{
Piyush Tiwary$^{1,3,}$\footnote{Work done as a Student Researcher at Google DeepMind.}
\and
Utkarsh Ahuja$^2$\and
Depanshu Sani$^1$\and
Aishwarya Jayagopal$^1$\and
Sagar Gubbi$^1$\and
Subhashini Venugopalan$^1$\and
Alok Talekar$^1$\And
Vaibhav Rajan$^1$
\affiliations
$^1$Google DeepMind, 
$^2$Google,
$^3$Indian Institute of Science\\
\emails
piyushtiwary@iisc.ac.in,\\
\{utkarshahuja, depanshus, ajayagopal, gubbi, vsubhashini, atalekar, vaibhavrajan\}@google.com
}
\begin{document}

\maketitle

\begin{abstract}
Agricultural landscape segmentation in the Global South is challenging as it is characterized by fragmented plots, high intra-class variance, and a scarcity of labeled training data. Recent advances in segmentation have been made by Multimodal Large Language Models (MLLMs). However, current approaches encounter critical context length bottlenecks and a domain alignment gap in understanding satellite features. We address these limitations through MAgSeg, a novel, decoder-free MLLM segmentation approach. MAgSeg is an architecturally efficient approach that enables standard MLLMs to perform segmentation of complex smallholder agricultural landscapes from high-resolution satellite imagery, without requiring auxiliary vision decoders. We introduce a novel instruction tuning data format designed to enable scalable fine-tuning and post-training on high resolution satellite imagery, which enables MAgSeg to learn from the global context of the image while generating text tokens for only a patch within the image. Extensive evaluations on datasets spanning three countries in the Global South demonstrate that MAgSeg significantly outperforms state-of-the-art MLLM baselines, offering a scalable solution to map smallholder agricultural environments.

\end{abstract}

\section{Introduction}
\label{sec:intro}

Agricultural Landscape Segmentation involves identifying the polygonal boundaries and constituent areas of agricultural features, such as crop fields, trees and water bodies, using satellite imagery.
It is a foundational step for large-scale mapping and analysis of agricultural environments \cite{fao2023state,oecd2023agricultural},
%Addressing the interconnected global crises of food insecurity, climate change, and resource degradation necessitates a comprehensive understanding of agricultural environments through large-scale mapping and analysis \cite{fao2023state,oecd2023agricultural}. 
%A foundational step in this direction is the precise delineation of crop fields\cite{Ma2019Deep}. 
which, in turn, support advanced, field-level analytics for multiple downstream precision agriculture applications such as crop type and yield management, policy making and resource allocation
\cite{rudel2009agricultural,Kang2019FieldLevel}. 
Effective solutions in these applications can have significant impact on mitigating climate change and managing global food security \cite{fao2023state,oecd2023agricultural}.
%However, a holistic view must also encompass landscape features beyond crop fields. 
%The delineation of trees and agroforestry systems, for instance, is vital for enhancing both agricultural sustainability and resilience \cite{pancholi2023role}, allowing for the quantification of their role in carbon sequestration and guiding the implementation of climate-resilient practices \cite{FAOSTAT_ClimateChange2023,Gao2023GreenhouseGE,lasco2014agroforestry}. In parallel, identifying and monitoring water resources, including on-farm ponds and wells, plays an indispensable role in sustainable water management by facilitating efficient irrigation planning \cite{cofie2015water,magidi2021informing}. 
%Ultimately, a granular, multi-faceted map that integrates fields, trees, and water systems provides the foundational data layer needed to foster intelligent policy, empower farmers, and build a sustainable food system.
This task is especially challenging for \textit{smallholder farms} in the Global South, which are characterized by an intricate mosaic of very small fields ($< 2$ ha) \cite{shdefinition}, diverse land uses, and interspersion with natural vegetation and water resources \cite{Lesiv2019Estimating,Samberg2016Subnational,Rada2019New}, and are responsible for more than half of the 
global agricultural production \cite{Samberg2016Subnational,sylvester2015success}.
%, and continue to face the problem of large yield gaps \cite{Mueller2012Closing}.
% containing fields smaller than 2 hectares \citep{fao2005economic}; and is crucial for the Global South comprising about two-thirds of the developing world’s 3 billion rural people in about 475 million %small farm
% households \citep{inbook}.

% The transition to a sustainable global food system hinges on our ability to digitize and monitor agricultural landscapes with high precision. Such analysis serves as the cornerstone of agricultural Digital Public Infrastructure (DPI), facilitating critical interventions ranging from crop yield forecasting to targeted farm subsidies. However, true sustainability requires looking beyond the crop field. Detailed mapping of agroforestry features is vital for calculating carbon sequestration potential and bolstering resilience against extreme heat and drought. Equally critical is the granular identification of hydrological structures, such as on-farm ponds and wells, which facilitate efficient irrigation planning and water quality protection. By delivering a holistic segmentation of fields, vegetation, and water bodies, we establish the ground-truth data required to drive smart policy and secure farmer livelihoods in an increasingly volatile climate.

The nature of smallholder systems
pose unique technical challenges in  landscape segmentation.
These include high variance within the \textit{field} class (due to diverse crops, shapes, and sizes), and low variance between fields and other classes like forests; and large distribution shifts with varying  landscape characteristics across different regions, climates, and seasons \cite{rs14225738,KernerEtAl2023Multi}.
Accurate delineation of small-sized fields and other features necessitate the use of
very-high-resolution (VHR) satellite imagery (e.g.,  WorldView-3 data at 0.31 to 2 m/pixel).
However, adapting standard image segmentation models, primarily developed for natural images, for this task remains challenging due to the high scale variation, densely packed small-scale objects and computational challenges arising from high dimensional inputs \cite{zhu2017deep,rolf2024mission}.
% While machine learning (ML) combined with high-resolution remote sensing offers a scalable pathway to generate these maps, applying standard computer vision techniques to smallholder farms presents unique challenges.
% These challenges stem primarily from the inherent 
% Additional technical difficulties stem from the nature of smallholder systems, 
% such as 
Moreover, nearly all previous approaches use specialized supervised learning architectures and rely on labelled data.
Obtaining such labeled data is %exceptionally 
labor-intensive, costly, and difficult to scale across the diverse geographies of the Global South.

% Recent works have tackled the smallholder field boundary delineation problem by modeling it as a semantic segmentation task, i.e., per-pixel binary classification where the classes are determined based on either boundary or extent of the field, e.g., \cite{mei2022using}.
% To achieve the final goal of delineating each field as a distinct object (i.e., instance segmentation), approaches have either relied on post-processing steps (like watershed algorithm \cite{vincent1991watersheds}) after semantic segmentation or employed direct object detection methods (e.g., Mask-RCNN \cite{he2017mask}). Further, to address the problems of insufficient labeled data and distributional shift across geographies, transfer learning has also been utilized \cite{KernerEtAl2023Multi,rs14225738}.
% All these works use specialized model architectures for image segmentation. 
% However, the recent paradigm shift toward Multimodal Large Language Models (MLLMs) -- such as GPT-4, Gemini, and Claude --raises a compelling question that has not yet been systematically investigated: \textit{Can we leverage the reasoning and generalization capabilities of these ubiquitous foundation models to directly map agricultural features?}

Multimodal Large Language Models (MLLMs) harness the inherent reasoning capabilities and broad world knowledge of LLMs, and have been recently used for various segmentation tasks with natural images \cite{zhou2024image}.
%Unifying agricultural mapping within a single general-purpose 
MLLMs offer the potential for scalability and accessibility with relatively lesser labeled data requirements to address the smallholder landscape segmentation problem. 
However, existing MLLM-based segmentation approaches perform poorly on VHR satellite images, particularly on this challenging task.
%face significant architectural hurdles. 
The dominant `embedding-as-mask' paradigm, exemplified by LISA~\cite{lisa} and GSVA~\cite{gsva}, relies on \textit{auxiliary decoder heads} to generate pixel-level masks. This design introduces parameter overhead and creates an alignment bottleneck between the LLM's text-centric latent space and the pixel decoder. %Alternative methods attempt to bypass decoders by providing polygon coordinates as outputs directly~\cite{visionllm,visionllm2}, but these models often struggle to grasp complex geometries, leading to topological errors. A more promising direction, 
The other `text-as-mask' paradigm is a \textit{decoder-free} approach
proposed by Text4Seg~\cite{text4seg}.
It encodes masks as text sequences
%using Run-Length Encoding (RLE), 
and seamlessly integrates segmentation into the standard autoregressive generation process of MLLMs.
%Despite its elegance, the text-based RLE approach 

However, this approach also fails when applied to VHR satellite imagery for two reasons. 
%The first reason is the context length bottleneck:
Encoding a full VHR mask (e.g.,  $615\times615$ px) would require about $\mathcal{O}(3.6 \times 10^5)$ tokens, far exceeding the context windows of many MLLMs.
Text4Seg relies on severe downsampling (e.g., to $32\times32$ px) to keep token counts of the encoded masks low.
While acceptable for natural images, this destroys critical high-frequency details in satellite data required for boundary delineation. 
The second reason is the domain alignment gap: general-purpose MLLMs lack intrinsic understanding of overhead satellite imagery. 
E.g., while they can identify `trees' in a photograph, they struggle to distinguish %spectrally
similar features like `crops' versus `pasture' in satellite views without explicit pixel-level grounding~\cite{lisat}.

In this work, we present a decoder-free \underline{M}LLM-based \underline{Ag}ricultural Landscape \underline{Seg}mentation approach (MAgSeg) that systematically overcomes these limitations.
We create an instruction tuning dataset with a novel format, which is designed to enable efficient fine-tuning and aims at retaining both local or global context in VHR masks.
This resolves the token length problem through a patch-based  strategy: MAgSeg receives the full VHR image as context but is tasked with generating a text-based mask for only a specific sub-patch. % thereby controlling its length.
Supervised fine tuning on this data helps, but is insufficient since it is optimized for next-token prediction (for the text-based mask) and not spatial accuracy.
% We propose a two-stage training strategy designed to adapt MLLMs for high-precision satellite segmentation without incurring massive computational costs. In the first stage, we employ LoRA-based Supervised Finetuning (SFT). To solve the token length issue, we introduce a patch-based prompting strategy: the model receives the full HR image as context but is tasked with generating the RLE mask for only a specific sub-patch. This preserves local detail while keeping sequence lengths manageable.
%In the second stage, 
We address this domain alignment gap via Group Relative Policy Optimization (GRPO)
%. Unlike standard SFT which optimizes for syntax (next-token prediction), GRPO uses Reinforcement Learning to optimize for semantics, 
using the mean-DICE score as a direct reward signal. This enables MAgSeg to prioritize pixel-level accuracy over mere textual fluency on the encoded masks.

We use another metric, \textit{overhead}, to measure the parameter efficiency of MLLM-based segmentation models which quantifies the additional \textit{learnable} parameters (e.g., in pixel decoders and segmentation heads) beyond the base LLM backbone; and serves as a proxy for computational cost and memory footprint during inference.
MAgSeg, which has zero overhead, shows dramatic gains in segmentation performance over previous zero-overhead decoder-free approaches in our experiments. Moreover, MAgSeg also significantly outperforms previous decoder-based methods which achieve better performance, relative to previous zero-overhead methods, at the cost of high overhead.

Our contributions are summarized as follows:
\begin{enumerate}[noitemsep,topsep=0pt,labelindent=0em,leftmargin=*]
    \item We propose an architecturally efficient, zero-overhead approach, MAgSeg, that enables standard MLLMs to perform segmentation of complex smallholder agricultural landscapes from very-high-resolution (VHR) satellite imagery, without requiring auxiliary vision decoders.
    %or complex polygon heads.
    \item MAgSeg overcomes critical context length bottlenecks and domain alignment gaps in previous decoder-free approaches through instruction tuning with a novel format, tasked to generate text tokens for only a patch within a VHR image, and RL-based post-training to refine pixel-level understanding. 
    %training pipeline combining patch-aware LoRA finetuning (to handle high-resolution inputs efficiently) with GRPO post-training (to align the model with pixel-level segmentation metrics).
    \item On challenging benchmark smallholder agriculture datasets spanning three countries in the Global South, MAgSeg significantly outperforms state-of-the-art LLM-based baselines. 
    %and achieves competitive parity with specialized, domain-specific vision models, demonstrating that general-purpose MLLMs can be effectively grounded for such scientific tasks.
\end{enumerate}

\section{Background and Related Works}
\label{sec:rel-works}

\textbf{MLLMs for Vision Tasks.} MLLMs 
%such as GPT-4, Gemini, and open-source variants like LLaVA~\cite{llava} and Gemma~\cite{gemma} 
%has revolutionized vision-language understanding. These models 
demonstrate exceptional capabilities in visual reasoning, captioning, and visual question answering. 
%Despite their success in high-level reasoning, 
However, applying MLLMs to dense prediction tasks like segmentation remains non-trivial due to the inherent modality gap between language and vision.
A common approach, exemplified by LISA~\cite{lisa} and GLaMM~\cite{glamm}, involves appending a specialized pixel decoder (e.g., SAM~\cite{sam}) to the LLM. The LLM outputs a special token (e.g., \texttt{<seg>}) which cues the decoder to generate a mask. While effective, this `embedding-as-mask' approach increases training complexity, limits model scalability
%introduces significant architectural complexity 
and disconnects the segmentation process from the LLM's native text-generation capabilities. 
Other approaches like VisionLLM~\cite{visionllm,visionllm2}  output polygon coordinates directly, but often struggle with complex geometries in satellite imagery. %Our method departs from these decoder-dependent architectures by treating segmentation as a pure text-generation task, seamlessly integrating it into the autoregressive paradigm of the LLM.

%\textbf{Instruction Tuning for Segmentation.} 
To address these limitations 
Text4Seg~\cite{text4seg} introduced
a decoder-free `text-as-mask' approach by casting segmentation as a sequence-to-sequence problem, which aligns with the autoregressive training pipeline of LLMs and enables scalable, flexible modeling with minimal architectural changes.
The key idea is to split the 
mask into a grid of fixed-size patches, flatten the patches and represent each patch with a \textit{semantic descriptor}, which can be a word, phrase or a more complex textual description.
Thus, we obtain a sequence of text tokens whose length can become exceedingly large with both number of patches and more complex descriptions.
To compress this description, they apply Run-Length Encoding (RLE) to the sequence, and empirically find that applying RLE row-wise (RRLE), with each row separated by a newline character, to the mask patches works well in practice.
This transformed visual instruction data is used to fine tune the MLLM.
% as discrete text sequences thereby
% and unify segmentation with language modeling, recent works have explored decoder-free approaches by representing masks as discrete text sequences. Text4Seg~\cite{text4seg} introduced ``semantic descriptors'' and Row Run-Length Encoding (RRLE) to cast segmentation as a sequence-to-sequence problem, achieving competitive results on natural images without auxiliary decoders. 

However, Text4Seg, which achieves state-of-the-art performance on several segmentation tasks on natural images, fails on the challenging task of smallholder landscape segmentation (as seen in our experiments).
%Detailed instance masks from complex agricultural 
These landscapes have both large number of objects to be delineated and high scale variation, e.g., ranging from small to very large fields.
Hence, the masks lead to complex and long descriptions with long sequences even after RRLE.
The approach of downsampling the masks used by \cite{text4seg} leads to loss of critical high-frequency spatial details required for boundary delineation, whereas
increasing the number of patches %to address this problem 
leads to loss of contextual information for each instance, which, in turn, deteriorates segmentation performance.
Our approach, MAgSeg, specifically addresses these problems, while retaining the advantages of their decoder-free approach.
% To address these problems, we develop a new format for creating the visual instruction dataset which can enable efficient fine tuning on HR images, without losing local or global context for each instance.
% % However, applying this ``text-as-mask'' paradigm to high-resolution (HR) satellite imagery presents two distinct challenges: the excessive token length required for detailed masks and the domain gap between natural and satellite imagery. While Text4Seg addresses token length via row-wise compression (R-RLE) and downsampling, this often leads to a loss of high-frequency spatial details critical for precision agriculture. 
% Furthermore, since supervised finetuning (SFT) optimizes for textual next-token likelihood rather than spatial accuracy, it fails to bridge the domain gap between the semantic descriptor sequence of a patch and pixel-level semantics.
% We solve this through Reinforcement Learning (RL) based post-training
% %We address these limitations by introducing a two-stage training framework. We leverage LoRA-based SFT~\cite{lora} for syntactic alignment and, crucially, incorporate Group Relative Policy Optimization (GRPO)~\cite{grpo} 
% to explicitly optimize pixel-level rewards. %This allows our model to learn robust, pixel-perfect segmentation for HR satellite imagery within a unified, decoder-free architecture.

\textbf{Agricultural Landscape Understanding.} 
% Remote sensing has long been a cornerstone for monitoring agricultural landscapes, yet smallholder systems present unique challenges due to fragmented plots, intercropped vegetation, and spectral ambiguity~\cite{alu}. 
Segmentation of agricultural landscapes has been extensively studied in remote sensing -- see \cite{zheng2025comprehensive} for a recent survey.
Most previous works have focussed on field boundary delineation, particularly in the global North, e.g., \cite{Masoud2020Delineation,waldner2020deep,waldner2021detect}.
A few recent works have recognized and addressed the challenges of smallholder agriculture by adapting classical computer vision techniques such as
panoptic segmentation \cite{alu},
object detection \cite{mei2022using}
and
transfer learning based methods \cite{rs14225738,KernerEtAl2023Multi}.
All these methods rely on supervised learning (mainly CNN-based) architectures.

To harness their robust reasoning capabilities, visual understanding and world knowledge, MLLMs
are being actively explored for various remote sensing tasks such as visual question answering and retrieval \cite{li2024vision}.
However, their use in segmentation tasks is relatively underexplored.
Two recent segmentation models, 
LISAt \cite{lisat} and 
FSVLM \cite{wu2025fsvlm} 
follow the same 
embedding-as-mask architecture of LISA~\cite{lisa}: a CLIP-encoded image \cite{radford2021learning} and a text embedding are passed to a MLLM (both use LLaVA~\cite{llava});
segmentation is triggered by a dedicated \texttt{<seg>} token within the LLM's vocabulary; 
upon generation, the token's embedding is mapped to a SAM-based decoder through an MLP; 
the decoder subsequently combines this embedding with the base image features to reconstruct the pixel-level mask.
The model is trained end-to-end with loss functions to optimize text generation and segmentation.
FSVLM was specifically evaluated for farmland segmentation while
LISAt  
demonstrated improvements over many geo-spatial foundation models on visual question answering, visual grounding and image captioning tasks.
In our experiments, we demonstrate significant improvements over this state-of-the-art approach on smallholder landscape segmentation datasets.

% Early approaches primarily relied on pixel-based semantic segmentation using standard CNN architectures like U-Net~\cite{unet} or DeepLab~\cite{deeplab}, often struggling to delineate smallholder field boundaries in high-resolution imagery. 
% More recent work has shifted toward instance segmentation to better capture field geometries. For example, \cite{maskrcnn_agri} adapted Mask-RCNN for field delineation in India, while \cite{resunet_a} and \cite{decode} introduced specialized multi-task architectures like ResUNet-a to predict field extent and boundaries simultaneously. 
% Transfer learning based techniques have utilized a combination of moderate and high resolution satellite imagery to combat both problems of insufficient labelled data in smallholder regions and  
% differences in farm sizes across regions \cite{rs14225738,KernerEtAl2023Multi}. 
% These methods largely ignore other critical features like agroforestry (trees) and water bodies (ponds, wells), which are essential for holistic landscape analysis~\cite{alu}. 
% % Our work builds upon the comprehensive multi-class framework introduced in the ALU system~\cite{alu}, which targets fields, trees, and water structures. 
% However, unlike ALU, which relies on complex post-processing heuristics and traditional CNN-based panoptic segmentation, we propose a generative approach leveraging the reasoning capabilities of Multimodal LLMs to perform segmentation directly from instruction prompts.

\section{Our Approach: MAgSeg}
\label{sec:methodology}
%\subsection{Overview}
%\label{sec:overview}
%We propose a decoder-free approach using multimodal LLMs to segment agricultural features—such as crop fields, trees, and ponds—from satellite imagery. 
Let $\gD = \{\rvx_i, \rvy_i\}_{i=1}^{N}$ 
denote a (training) dataset, where $\rvx_i \in \R^{h_\rvx \times w_\rvx \times 3}$ represents a high-resolution satellite RGB image with height $h_x$, width $w_x$, 3 channels, 
%derived from S2 cells, 
and $\rvy_i$ is the corresponding multi-class instance map. 
Our aim is to train a base MLLM, $\pi_\theta$, with parameters $\theta$,
using $\gD$, for the task of instance segmentation.

\begin{figure*}[h]
    \centering
    % Replace 'method_diagram.png' with your actual filename
    \includegraphics[width=0.9\linewidth]{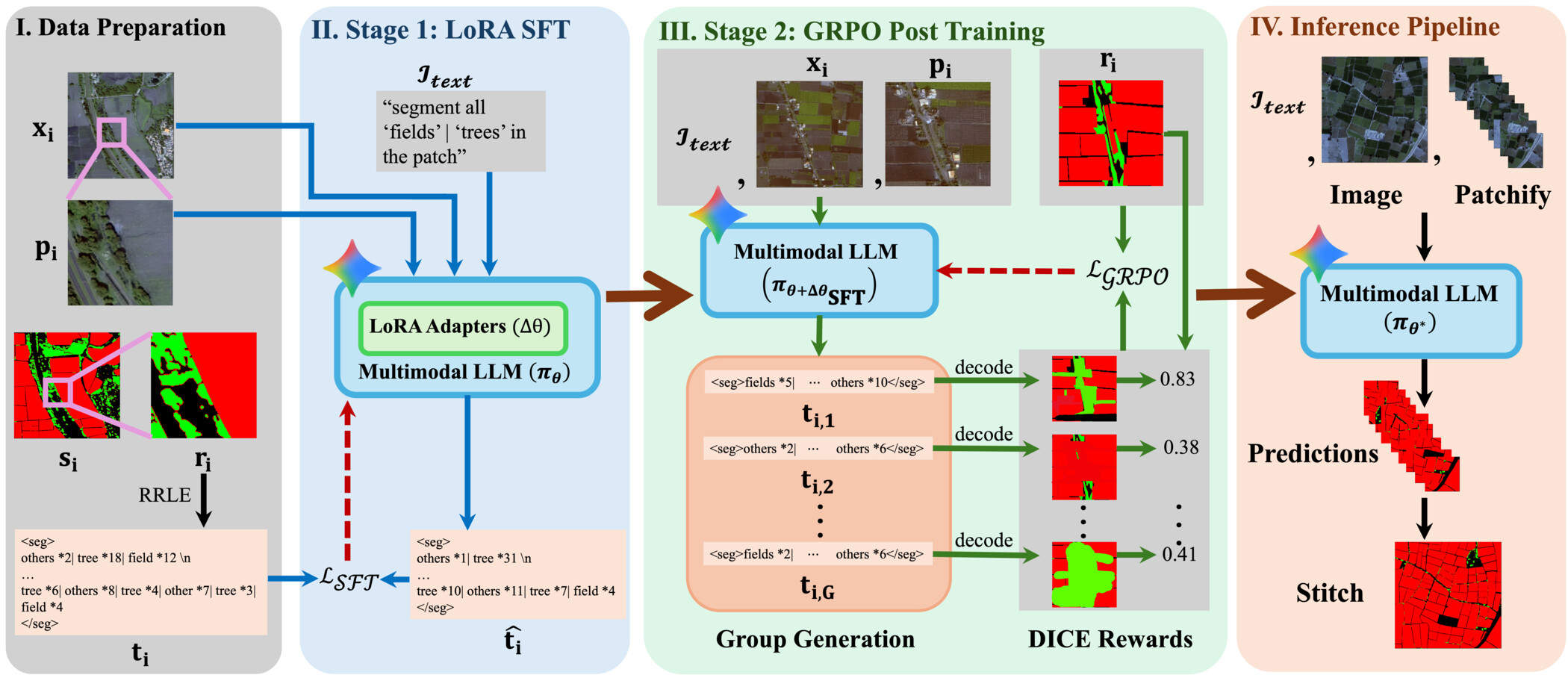} 
    \caption{\textbf{Overview of MAgSeg.} 
    \textbf{Data Preparation}: from each high-resolution satellite image $\rvx_i$ and its segmentation map $\rvs_i$, multiple patches $\rvp_i$ and their corresponding masks $\rvr_i$ are extracted, the masks are converted to a text-based RRLE representation $\mathbf{t}_i$ to form the instruction tuning dataset: $\{ \mathcal{I}_{\text{text}}, \rvx_i, \rvp_i \} \rightarrow \mathbf{t}_i$.
    \textbf{Training}: consists of two stages: (1) LoRA Supervised Finetuning (SFT), where the base multimodal LLM is adapted to generate RRLE-encoded masks using a frozen backbone and trainable LoRA adapters; and (2) GRPO Post-Training, where we sample multiple candidate RLE sequences, decode them into segmentation masks, and use the mean-DICE score as a reward signal to update the policy, ensuring pixel-perfect alignment.
    \textbf{Inference}: uses the trained MLLM to predict on a grid of disjoint patches, which are then stitched and postprocessed to form the final segmentation map.}
    \label{fig:method_overview}
    \vspace{-10pt} % Adjust this negative space to tighten the layout if needed
\end{figure*}

We first transform $\gD$ to an instruction tuning dataset with a novel format, designed to enable efficient fine-tuning without losing  local or global context in the high-resolution images.
Such supervised fine tuning (SFT), which has demonstrated success with natural images, proves insufficient for complex satellite imagery as it lacks the fine-grained spatial reasoning required for pixel-perfect segmentation.
Thus, we employ a Reinforcement Learning (RL) based post-training optimization step to maximize pixel-level semantic alignment.
The trained model is used to generate semantic masks which are post processed to form the instance masks during inference.
Figure \ref{fig:method_overview} has an overview of MAgSeg.
% The proposed methodology proceeds in two stages. First, we employ LoRA supervised-finetuning (SFT) on a base model, $\pi_\theta$, training it to generate Row Run-Length Encoding (RRLE) masks for specific image patches. This yields the finetuned parameter set $\pi_{\theta+\Delta\theta}$. In the second stage, we implement GRPO post-training to further refine the model. By utilizing the generated RLE mask to compute the mean-DICE score as a reward signal, we encourage the model to maximize pixel-level semantic alignment. This post-training step significantly improves the model's grasp of nuanced satellite imagery, a capability often lacking in naive finetuning approaches.

\subsection{Confronting High Resolution: Patch-based Instruction Tuning with Global Context}
\label{sec:stage_1}

Given a dataset sample $(\rvx_i, \rvy_i)$, we first derive the semantic segmentation map, $\rvs_i$ from the instance map $\rvy_i$. To facilitate patch-based processing, we extract a random crop from the high-resolution input. We sample top-left coordinates $(j, k)$ uniformly such that $j \sim \mathcal{U}(1, h_\rvx - h_\rvp)$ and $k \sim \mathcal{U}(1, w_\rvx - w_\rvp)$. Thus, from $\rvs_i$, this yields an image patch $\rvp_i$ of dimensions $h_\rvp \times w_\rvp$ and a spatially corresponding semantic mask patch $\rvr_i$. 

The mask patch $\rvr_i$ is subsequently converted into a text-based RRLE representation, denoted as $\mathbf{t}_i = \texttt{RRLE}(\rvr_i)$, following the semantic descriptors devised for Text4Seg~\cite{text4seg} (cf. Algorithm~\ref{alg:rle_encoding} in Appendix \ref{sec:rrle_algorithm}). We then construct an instruction-tuning sample consisting of the prompt input and the target response:
$$\text{Input:} \{ \mathcal{I}_{\text{text}}, \rvx_i, \rvp_i \} \quad \rightarrow \quad \text{Target:} \mathbf{t}_i$$
Here, 
%$\rvx_i$ serves as the global context image, $\rvp_i$ is the specific target patch for segmentation, and 
$\mathcal{I}_{\text{text}}$ is the textual prompt instructing the LLM to generate the RRLE mask for the provided patch.
Note that the entire high-resolution image $\rvx_i$ is passed with every patch $\rvp_i$ which provides global context for each sample pair.
Since the target $\mathbf{t}_i$
contains tokens only for the patch $\rvp_i$, we stay within the context window of the MLLM model. This format enables us to leverage the decoder-free seq-to-seq framework of Text4Seg on HR satellite images, without downsampling the entire HR image and losing critical high frequency details.

To fine-tune $\pi_\theta$ we employ Low-Rank Adaptation (LoRA)~\cite{lora} in conjunction with a standard autoregressive objective. 
LoRA optimizes a significantly smaller set of parameters, denoted as $\Delta\theta$, by freezing $\theta$ and injecting trainable rank decomposition matrices into the linear layer of the MLLM's transformer. This approach ensures scalability, data efficiency as well as generalizability.

Formally, let the ground-truth RRLE sequence be represented as a series of tokens $\mathbf{t}_i = \{w_1, w_2, \dots, w_T\}$. The trainable parameters $\Delta\theta$ are optimized by minimizing the negative log-likelihood of these target tokens conditioned on the multimodal context-
\begin{align}
&\Delta\theta_{\text{SFT}} = \arg\min_{\Delta\theta} \gL_{\text{SFT}}(\Delta\theta)\\
\gL_{\text{SFT}}(\Delta\theta) &= - \sum_{t=1}^{T} \log \pi_{\theta + \Delta\theta}(w_t \mid w_{<t}, \rvx_i, \rvp_i, \mathcal{I}_{\text{text}})
\end{align}
where, $\pi_{\theta + \Delta\theta}(\cdot)$ denotes the probability distribution of the next token predicted by the adapted model. Minimizing this objective aligns the model's visual reasoning with the textual RRLE structure, enabling segmentation of the target patch $\rvp_i$.

% Rather than performing full-parameter finetuning — which is computationally expensive and prone to catastrophic forgetting — LoRA freezes the pre-trained model weights $\theta$ and injects trainable rank-decomposition matrices into the linear layers of the MLLM's Transformer. This formulation allows us to optimize a significantly smaller set of parameters, denoted as $\Delta\theta$, while preserving the generalization capabilities of the base model. This approach ensures the SFT process remains scalable and data-efficient.
%\subsection{LoRA Supervised Finetuning}
%The initial stage of our framework involves supervised finetuning (SFT) of the base LLM using low-rank adaptation (LoRA)~\cite{lora}. 
After this fine-tuning stage, the MLLM is 
(a) conditioned to generate masks in the specified RRLE format, and
(b) instilled with a foundational understanding of segmentation semantics and class labels, as established in \cite{text4seg}.
% The primary objectives of this phase are twofold: (a) to condition the model to generate masks in the specified RRLE format, and (b) to instill a foundational understanding of segmentation semantics and class labels, as established in prior work~\cite{text4seg}.

% While standard SFT has demonstrated success with natural images, they often prove insufficient for complex satellite imagery. We hypothesize that this performance gap stems from the model's limited pre-training exposure to such images. 
% Specifically, while SFT successfully conditions the model to generate the correct RRLE syntax, it lacks the fine-grained spatial reasoning required for pixel-perfect segmentation.
% To bridge this gap, we propose a second stage of Reinforcement Learning (RL) post-training designed to explicitly optimize for pixel-level semantic accuracy.

\subsection{Mitigating Domain Gap with GRPO}
\label{sec:stage_2}
% The SFT phase provides a model capable of generating RRLE mask in appropriate format with basic segmentation capabilities. However, it lacks reasoning over the target patch with context image to generate pixel-perfect mask. Hence, to improve the model's segmentation capabilities at fine-grained pixel-level details, we use RL post-training. In particular, we use Group Relative Policy Optimization (GRPO)~\cite{grpo} to optimize the LLM / policy model, $\pi_\theta$ to maximize the pixel-level reward for the generated RRLE mask. To do this, for a given input $\{ \mathcal{I}_{\text{text}}, \rvx_i, \rvp_i \}$ we generate $B$ responses, $\{\rvt_i\}_{i=1}^{B}$ from the policy model. Each of these reponses are different instances of RRLE mask for the target patch, $\rvp_i$. We decode these masks back into 2D semantic arrays denoted by $\texttt{decode\_RRLE}(\rvt_i) = \Tilde{\rvs}_i$. 
% While the SFT phase aligns the model with the required RRLE syntax, it 
The previous SFT step 
primarily optimizes for \textit{next-token prediction} rather than \textit{spatial accuracy}. Consequently, the model often struggles 
%with fine-grained spatial reasoning, failing 
to achieve pixel-perfect alignment between the target patch and the context image. To bridge this gap, we employ Group Relative Policy Optimization (GRPO)~\cite{grpo} as post training step, where the optimization objective is shifted from maximizing likelihood to maximizing a pixel-level semantic reward.

For a given input tuple $\{ \mathcal{I}_{\text{text}}, \rvx_i, \rvp_i \}$, we sample a group of $G$ distinct outputs from the policy model $\pi_\theta$, denoted as $\{\mathbf{t}_{i,g}\}_{g=1}^{G}$. Each output $\mathbf{t}_{i,g}$ represents a candidate RRLE sequence for the target patch $\rvp_i$. To evaluate the quality of these candidates, we first decode them into 2D semantic masks:
\begin{align}
\tilde{\rvs}_{i,g} = \texttt{decode\_RRLE}(\mathbf{t}_{i,g})
\end{align}
where, $\tilde{\rvs}_{i,g}$ is the predicted segmentation map for the $g$-th sample (cf. Algorithm~\ref{alg:rle_decoding} in Appendix \ref{sec:rrle_algorithm}). 
To quantify the semantic accuracy, we compute the reward $r_{i,g}$ as the Mean DICE score (mDICE) between the predicted mask $\tilde{\rvs}_{i,g}$ and the ground truth mask $\rvs_i$. Let $C$ denote the set of classes (e.g., fields, ponds, trees). The reward is defined as:
\begin{align}
r_{i,g} = \frac{1}{|C|} \sum_{c \in C} \frac{2 \sum_{x,y} (\tilde{\rvs}_{i,g}^{(c)} \cap \rvs_i^{(c)})}{\sum_{x,y} \tilde{\rvs}_{i,g}^{(c)} + \sum_{x,y} \rvs_i^{(c)} + \epsilon}
\label{eq:dice_reward}
\end{align}
where, $\cap$ denotes the intersection of active pixels for class $c$, and $\epsilon$ is a smoothing term. GRPO optimizes the policy by estimating the advantage of each sample relative to the group average, eliminating the need for a separate value function critic. The advantage $A_{i,g}$ for the $g$-th sample is computed as:
\begin{align}
A_{i,g} = \frac{r_{i,g} - \mu_{r}}{\sigma_{r}}
\end{align}
where, $\mu_r$ and $\sigma_r$ are the mean and standard deviation of the rewards within the group of $G$ samples. The GRPO loss objective is formulated as:
% \begin{equation}
% \begin{split}
% \mathcal{L}_{\text{GRPO}}(\theta) = - \frac{1}{G} \sum_{g=1}^{G} \min \Bigg( & \frac{\pi_\theta(\mathbf{t}_{i,g})}{\pi_{\text{old}}(\mathbf{t}_{i,g})} A_{i,g}, \\
% & \text{clip}\left(\frac{\pi_\theta(\mathbf{t}_{i,g})}{\pi_{\text{old}}(\mathbf{t}_{i,g})}, 1-\epsilon, 1+\epsilon\right) A_{i,g} \Bigg)
% \end{split}
% \end{equation}
% \begin{align}
% % \small
% \mathcal{L}_{\text{GRPO}}(\theta) = - \frac{1}{G} \sum_{g=1}^{G} \min \left( \frac{\pi_\theta(\mathbf{t}_{i,g})}{\pi_{\text{old}}(\mathbf{t}_{i,g})} A_{i,g}, \text{clip}\left(\frac{\pi_\theta(\mathbf{t}_{i,g})}{\pi_{\text{old}}(\mathbf{t}_{i,g})}, 1-\epsilon, 1+\epsilon\right) A_{i,g} \right)
% \end{align}
% \begin{equation}
% \begin{split}
%     \mathcal{L}_{\text{GRPO}}(\theta) = - \frac{1}{G} \sum_{g=1}^{G} \min \Big( & \rho_{i,g} A_{i,g}, \\
%     & \text{clip}(\rho_{i,g}, 1-\epsilon, 1+\epsilon) A_{i,g} \Big)
% \end{split}
% \end{equation}
%\begin{align}
{\footnotesize
\begin{align}
    \mathcal{L}_{\text{GRPO}}(\Delta\theta) = - \frac{1}{G} \sum_{g=1}^{G} \min \Big( \rho_{i,g} A_{i,g}, 
     \text{clip}(\rho_{i,g}, 1-\epsilon, 1+\epsilon) A_{i,g} \Big)
\label{eq:grpo_loss}
\end{align}}
%\end{align}
\noindent where $\rho_{i,g} = \frac{\pi_{\theta+\Delta\theta}(\mathbf{t}_{i,g})}{\pi_{\theta+\Delta\theta_{\text{SFT}}}(\mathbf{t}_{i,g})}$ is the probability ratio. The final post-training objective is the GRPO loss with KL penalty:
\begin{align}
    \mathcal{L}_{\text{PT}} = \mathcal{L}_{\text{GRPO}}(\Delta\theta) + \beta D_{\text{KL}}(\pi_{\theta+\Delta\theta_{\text{SFT}}} || \pi_{\theta+\Delta\theta}) 
\label{eq:pt_loss}
\end{align}
This objective encourages the model to increase the probability of generating RRLE sequences that yield higher DICE scores, thereby explicitly refining the model's pixel-level understanding via the GRPO loss while at the same time avoids diverging from the reference policy obtained from Stage I, $\pi_{\theta+\Delta\theta_{\text{SFT}}}$.

\subsection{Inference}
During inference, we patchify the high-resolution input image into a grid of disjoint spatial patches of dimensions $h_\rvp \times w_\rvp$ (cf. Inference pipeline in Figure~\ref{fig:method_overview}). These patches are collated into a single batch to enable parallel processing, which significantly reduces computational latency compared to sequential execution. Once the individual patch masks are generated, they are spatially concatenated to reconstruct the full-resolution semantic map. To remove stitching artifacts and improve boundary precision, we apply the postprocessing steps recommended in EPOC \cite{chen2024subobject}.
This includes (i) boundary detection using a SegFormer~\cite{xie2021segformer} based refinement model and (ii) the Watershed algorithm~\cite{vincent1991watersheds} on the refined semantic maps to extract discrete instance segmentation maps.
%-level features for evaluation against ground truth labels.

\section{Experiments}
\label{sec:experiments}

\subsection{Experiment Settings}

\noindent\textbf{Datasets.}
We conduct our experiments on two datasets which span across 3 countries in the Global South with smallholder agriculture systems.
The first is the 
%To validate the efficacy of our proposed method, we conduct extensive evaluations on the challenging small-holder 
ALU dataset~\cite{alu} which contains 2009/430/431 images in its train/val/test splits.
%high-resolution agricultural samples 
This data is from India, annotated with five distinct semantic classes: fields, trees, clouds, ponds, and wells.
The second is a publicly available benchmark dataset, AI4SmallFarms~\cite{ai4}, for agricultural field boundary delineation. %(i.e., single class instance segmentation).
The train/val/test splits contain 1962/90/304 images for Vietnam and 1860/605/177 images for Cambodia.
In both cases, VHR imagery from Google Maps Satellite View are used. We take images sampled at 0.5 m/pixel GSD, from two satellite sources: Maxar Worldview, Airbus Pleiades.

\noindent\textbf{Baselines.} We benchmark our approach against 
%two distinct categories of 
state-of-the-art segmentation MLLM-based methods: LISA~\cite{lisa}, GSVA~\cite{gsva}, which are decoder-dependent and the decoder-free Text4Seg~\cite{text4seg}. We consider two variants of Text4Seg -- coarse predictions (C) and refined predictions (R) that uses SAM refinement. Further, we also consider LISAt~\cite{lisat} which extends LISA for segmentation in satellite images.
In addition, we include Transformer-based models specialized for segmentation, LAVT~\cite{lavt} and GRES~\cite{gres}.
% \begin{itemize}
%     \item \textbf{Decoder-Dependent MLLMs:} We compare against Multimodal Large Language Models that rely on additional decoder heads for pixel-level generation, specifically LISA~\cite{lisa}, GSVA~\cite{gsva}, GRES~\cite{gres}. Further, we also consider LISAt~\cite{lisat} which is extends the design of LISA for segmentation in satellite images.
%     \item \textbf{Segmentation Architectures:} We also include pure Transformer and CNN-based vision-language models specialized in segmentation, namely LAVT~\cite{lavt} and the original ALU baseline~\cite{alu}.
% \end{itemize}

\noindent\textbf{Evaluation Metrics.} Following the evaluation protocol in \cite{alu} for smallholder farms, we report the instance-wise Mean Intersection-over-Union (mIoU), Median IoU, False Positive Rate (FPR), and False Negative Rate (FNR). More details are in Appendix \ref{sec:metrics_defn}.

In addition to standard performance metrics, we use a parameter efficiency metric, denoted as \textbf{Overhead}. This metric quantifies the additional learnable parameters required by a method beyond the base LLM backbone. For decoder-dependent baselines, this includes parameters of the external segmentation heads such as SAM-H, pixel decoders, and any projection layers. For decoder-free methods such as ours and Text4Seg, the overhead is zero. Concretely, total number of parameters of a method is denoted as -
\begin{align}
    \Phi_{\text{total}} = \Phi_{\text{base}} + \Phi_{\text{projection}} + \Phi_{\text{decoder}}
\label{eq:total_params}
\end{align}
where, $\Phi_{\text{base}}$, $\Phi_{\text{projection}}$ and $\Phi_{\text{decoder}}$ denote the parameters in the base model, projection layers and decoder heads respectively. Then, the overhead is defined as -
\begin{align}
    \text{Overhead} = \frac{\Phi_{\text{projection}} + \Phi_{\text{decoder}}}{\Phi_{\text{base}}}
\label{eq:overhead_metric}
\end{align}
This metric is vital for two reasons: (1) it highlights the efficacy of the base LLM's native reasoning capabilities without reliance on auxiliary decoder heads, and (2) it serves as a proxy for the computational cost and memory footprint required during inference\footnote{Here, we refer solely to the additional overhead incurred during training; inference overhead is excluded.}.

\noindent\textbf{Implementation.} For MAgSeg, we experiment with MLLMs of two sizes -- 4B and 12B -- versions of Gemma3~\cite{gemma}. We refer to these two variants as MAgSeg (4B) and MAgSeg (12B) respectively.
These are comparable to the model sizes used in the MLLM-based baseline methods: 
LISA (13B), LISAt (7B), GSVA (13B) and Text4Seg (7B).
We use $h_\rvp = w_\rvp = 32$ for patch dimension as mentioned in Section~\ref{sec:methodology}. 
Appendix \ref{sec:implementation} has other details of MAgSeg and all the baselines.

\subsection{Results}
\subsubsection{Segmentation Performance}
\label{sec:main-results}

\noindent\textit{India.}
The quantitative comparison on the panoptic segmentation dataset from India is summarized in Table~\ref{tab:main-result}. Our proposed method, MAgSeg, advances the state-of-the-art, outperforming all MLLM-based and transformer-based baselines by a substantial margin. Specifically, we observe an improvement of $21$ mIoU points over GRES, the closest competitor. 
%While the specialized ALU model~\cite{alu}—designed specifically for field boundary delineation—retains the top performance, our method ranks second overall, demonstrating that generalist LLMs can approach the performance of domain-specific experts without specialized encoders.

\noindent\textit{Vietnam and Cambodia.}
% To validate the capabilities of our framework across distinct agricultural topologies, we conducted independent training and evaluation on the AI4SmallFarms ataset~\cite{KernerEtAl2023Multi}, specifically the Vietnam and Cambodia subsets. 
These regions are characterized by dense wetland rice cultivation, offering a different spatial distribution compared to India. As shown in Table~\ref{tab:main-result-ai4sf}, MAgSeg consistently achieves state-of-the-art results when trained on these diverse landscapes.
The performance gap is distinct in Cambodia split. Here, baseline methods struggle to learn effective segmentation boundaries; notably, even strong decoder-based models like GRES and LISA plateau at approximately $0.12$ mean IoU. Text4Seg fails completely ($0.04$ mean IoU) due to the small, fragmented nature of agricultural fields which are lost during mask downsampling. In contrast, MAgSeg successfully learns the complex landscape features, achieving a mean IoU of $0.43$. This result -- nearly quadrupling the performance of the nearest baseline -- demonstrates that our GRPO-based training objective enables the model to converge on difficult pixel-level tasks where standard cross-entropy or decoder-alignment losses fail.
In the Vietnam data, while baseline performance improves (with GRES reaching $0.40$ mean IoU), our MAgSeg-12B model maintains better performance with a mean IoU of $0.42$ and a better median IoU of $0.35$. 
%Lastly, we also evaluate the field delineation performance using PoLiS score and boundary classification metrics like Precision, Recall and F1-Score. These results are provided in Table~\ref{tab:ai4sf-metrics} in Appendix.

\begin{table*}[!h]
\centering
\resizebox{0.85\textwidth}{!}{%
\begin{tabular}{l|l|rrrrrrr|rr}
\toprule
\multicolumn{1}{c|}{\textbf{Class}} & \multicolumn{1}{c|}{\textbf{Metric}} & \multicolumn{1}{c}{\textbf{LAVT}} & \multicolumn{1}{c}{\textbf{LISA}} & \multicolumn{1}{c}{\textbf{LISAt}} & \multicolumn{1}{c}{\textbf{GSVA}} & \multicolumn{1}{c}{\textbf{GRES}} & \multicolumn{1}{c}{\textbf{Text4Seg (C)}} & \multicolumn{1}{c|}{\textbf{Text4Seg (R)}} & \multicolumn{1}{c}{\begin{tabular}[c]{@{}c@{}}\textbf{MAgSeg (4B)}\\ (Our)\end{tabular}} & \multicolumn{1}{c}{\begin{tabular}[c]{@{}c@{}}\textbf{MAgSeg (12B)}\\ (Our)\end{tabular}} \\ \midrule  \midrule
%\multicolumn{2}{c|}
& \textbf{Overhead} & $9.95\%$ & $9.48\%$ & $9.48\%$ & $\underline{9.35}\%$ & $15.82\%$ & $\textbf{0.00}\%$ & $\textbf{0.00}\%$ & $\textbf{0.00}\%$ & $\textbf{0.00}\%$ \\ \midrule
\multirow{4}{*}{\textbf{Fields}} & \textbf{mean IoU}  & 0.22 & 0.22 & 0.36 & 0.35 & 0.37 & 0.05 & 0.07 & \underline{0.58}& \textbf{0.59}\\
 & \textbf{median IoU}  & 0.13 & 0.05 & 0.30 & 0.29 & 0.31 & 0.02 & 0.02 & \underline{0.62}& \textbf{0.64}\\
 & \textbf{FNR}  & 8.08 & 44.25 & 6.06 & 7.85 & \underline{3.48}& 17.24 & 20.36 & 4.71 & \textbf{3.13}\\
 & \textbf{FPR}  & 40.66 & 38.72 & 52.28 & 32.58 & 30.98 & 32.59 & 44.99 & \textbf{20.15}& \underline{21.22}\\ \midrule
\multirow{4}{*}{\textbf{Trees}} & \textbf{mean IoU}  & 0.11 & 0.00 & 0.11 & 0.03 & 0.13 & 0.00 & 0.00 & \underline{0.20}& \textbf{0.21}\\
 & \textbf{median IoU}  & 0.01 & 0.00 & 0.02 & 0.00 & 0.02 & 0.00 & 0.00 & \underline{0.09}& \textbf{0.10}\\
 & \textbf{FNR}  & 45.43 & 98.63 & 39.13 & 87.05 & 39.83 & 89.39 & 91.28 & \underline{37.26}& \textbf{34.22}\\
 & \textbf{FPR}  & 52.19 & 100.00 & 54.74 & 61.47 & \textbf{37.05}& 68.38 & 76.01 & \underline{42.08}& 42.89 \\ \midrule
\multirow{4}{*}{\textbf{Clouds}} & \textbf{mean IoU}  & \underline{0.19}& 0.00 & 0.07 & 0.03 & 0.14 & 0.08 & 0.09 & \textbf{0.25}& \textbf{0.25}\\
 & \textbf{median IoU}  & 0.03 & 0.00 & 0.00 & 0.00 & 0.00 & 0.00 & 0.00 & \underline{0.15}& \textbf{0.17}\\
 & \textbf{FNR}  & 24.53 & 100.00 & 66.04 & 86.79 & 47.17 & 37.74 & 45.28 & \underline{19.82}& \textbf{19.28}\\
 & \textbf{FPR}  & 56.10 & 100.00 & 43.12 & 91.04 & 42.47 & \textbf{29.57}& \underline{32.58}& 41.22 & 39.08 \\ \midrule
\multirow{4}{*}{\textbf{Ponds}} & \textbf{mean IoU}  & 0.00 & 0.00 & 0.00 & \underline{0.02}& 0.01 & 0.00 & 0.00 & \textbf{0.05}& \textbf{0.05}\\
 & \textbf{median IoU}  & 0.00 & 0.00 & 0.00 & 0.00 & 0.00 & 0.00 & 0.00 & 0.00 & 0.00 \\
 & \textbf{FNR}  & 100.00 & \underline{97.66}& 100.00 & \textbf{93.38}& 97.90 & 100.00 & 100.00 & 99.56 & 99.16 \\
 & \textbf{FPR}  & 100.00 & 100.00 & 100.00 & 99.47 & \textbf{68.42}& 100.00 & 100.00 & \underline{95.62} & 96.31 \\ \midrule
\multirow{4}{*}{\textbf{Well}} & \textbf{mean IoU}  & 0.00 & 0.00 & 0.00 & 0.00 & 0.00 & 0.00 & 0.00 & 0.00 & 0.00 \\
 & \textbf{median IoU}  & 0.00 & 0.00 & 0.00 & 0.00 & 0.00 & 0.00 & 0.00 & 0.00 & 0.00 \\
 & \textbf{FNR}  & 100.00 & 100.00 & 100.00 & 100.00 & 100.00 & 100.00 & 100.00 & 100.00 & 100.00 \\
 & \textbf{FPR}  & 100.00 & 100.00 & 100.00 & 100.00 & 100.00 & 100.00 & 100.00 & 100.00 & 100.00 \\ \bottomrule
\end{tabular}%
}
\caption{Quantitative results on the India (ALU) panoptic segmentation dataset. We report instance-wise mean IoU ($\uparrow$), median IoU ($\uparrow$), FPR ($\downarrow$), FNR ($\downarrow$), and model overhead ($\downarrow$).}
\label{tab:main-result}
\end{table*}

% Please add the following required packages to your document preamble:
% \usepackage{graphicx}
\begin{table*}[!h]
\centering
\resizebox{0.8\textwidth}{!}{%
\begin{tabular}{c|l|rrrrrrr|rr}
\toprule
\textbf{Country} & \multicolumn{1}{c|}{\textbf{Metric}} & \multicolumn{1}{c}{\textbf{LAVT}} & \multicolumn{1}{c}{\textbf{LISA}} & \multicolumn{1}{c}{\textbf{LISAt}} & \multicolumn{1}{c}{\textbf{GSVA}} & \multicolumn{1}{c}{\textbf{GRES}} & \multicolumn{1}{c}{\textbf{Text4Seg (C)}} & \multicolumn{1}{c|}{\textbf{Text4Seg (R)}} & \begin{tabular}[c]{@{}c@{}} \textbf{MAgSeg (4B)}\\ (Our)\end{tabular} & \multicolumn{1}{c}{\begin{tabular}[c]{@{}c@{}} \textbf{MAgSeg (12B)}\\ (Our)\end{tabular}} \\ \midrule
\multirow{4}{*}{\textbf{Cambodia}} & \textbf{mean IoU} & 0.06 & 0.09 & 0.12 & 0.10 & 0.12 & 0.04 & 0.04 & \underline{0.41} & \textbf{0.43} \\
 & \textbf{median IoU} & 0.02 & 0.03 & \underline{0.05} & 0.04 & \underline{0.05} & 0.02 & 0.01 & \textbf{0.45} & \textbf{0.45} \\
 & \textbf{FNR} & \underline{2.11} & 3.11 & 3.37 & 3.60 & \textbf{0.06} & 6.21 & 5.31 & 8.69 & 8.59 \\
 & \textbf{FPR} & 15.12 & 33.90 & 22.75 & 32.81 & \textbf{2.56} & 19.83 & 31.79 & 16.69 & \underline{14.66} \\ \midrule
\multirow{4}{*}{\textbf{Vietnam}} & \textbf{mean IoU} & 0.19 & 0.22 & 0.24 & 0.24 & \underline{0.40} & 0.08 & 0.09 & \underline{0.40} & \textbf{0.42} \\
 & \textbf{median IoU} & 0.07 & 0.10 & 0.13 & 0.12 & 0.31 & 0.03 & 0.03 & \underline{0.33} & \textbf{0.35} \\
 & \textbf{FNR} & 9.30 & 10.01 & 5.51 & 8.90 & \underline{3.67} & 11.02 & 11.45 & 3.99 & \textbf{3.52} \\
 & \textbf{FPR} & \underline{6.28} & 31.84 & 34.77 & 31.51 & 10.60 & 21.75 & 39.69 & 11.47 & \textbf{3.86} \\ \bottomrule
\end{tabular}%
}
\caption{Quantitative results on the AI4SmallFarms field segmentation dataset. We report instance-wise mean IoU ($\uparrow$), median IoU ($\uparrow$), FPR ($\downarrow$), and FNR ($\downarrow$) for the Vietnam and Cambodia subsets.}
\label{tab:main-result-ai4sf}
\end{table*}

\begin{figure*}[!h] % Use figure* for full-width in two-column papers
    \centering
    % Adjust the width if you want margins (e.g., 0.95\linewidth)
    \resizebox{0.9\textwidth}{!}{ 
        \input{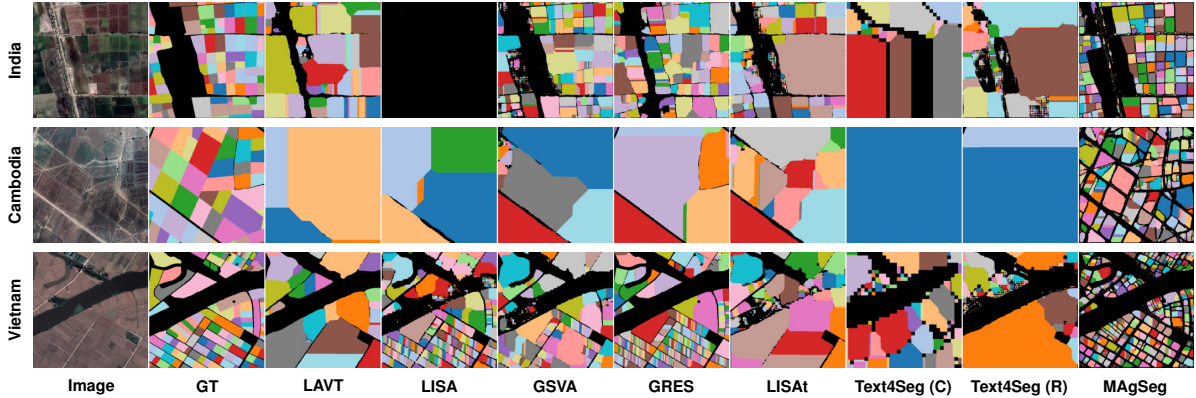}
    }
    \caption{Qualitative results comparing our approach MAgSeg with SOTA baselines. GT: Ground Truth.}
    \label{fig:combined_qualitative_results}
\end{figure*}

% \begin{table}[]
% \centering
% \resizebox{\columnwidth}{!}{%
% \begin{tabular}{c|rrrr}
% \toprule
% \textbf{Method} & \multicolumn{1}{c}{\textbf{Precision} $(\uparrow)$} & \multicolumn{1}{c}{\textbf{Recall} $(\uparrow)$} & \multicolumn{1}{c}{\textbf{F1-Score} $(\uparrow)$} & \multicolumn{1}{c}{\textbf{PoLiS (m)} $(\downarrow)$} \\ \midrule
% \textbf{GM} & 0.47 & 0.52 & 0.49 & 20.30 \\
% \textbf{Ours} & 0.32 & 0.91 & 0.42 & 7.11 \\ \bottomrule
% \end{tabular}%
% }
% \caption{Comparison of field boundary classification metrics on AI4SmallFarms dataset.}
% \label{tab:ai4sf-metrics}
% \end{table}

We attribute the significantly lower performance of prior methods to their inherent architectural limitations, which are addressed in MAgSeg: 
(a) \textit{Text4Seg:} This method exhibits poor performance on satellite imagery. As noted in \cite{text4seg}, Text4Seg relies on downsampling masks (e.g., to $32\times32$) to maintain tractable token counts. In the context of high-resolution satellite data, this compression causes a severe loss of high-frequency spatial details -- such as narrow field boundaries -- which cannot be recovered during inference. 
(b) \textit{Decoder-Based Models (LISA, GSVA, GRES):} While methods like GRES perform better than Text4Seg, they still struggle to match MAgSeg. Specifically, for field segmentation on the ALU dataset, our method outperforms the best decoder-based model (GRES) by $12$ and $33$ points in mIoU and median IoU, respectively. On the Cambodia dataset, the gains are even larger—$31$ and $40$ points in mIoU and median IoU. These substantial improvements validate our claim.
We attribute this to the `domain alignment gap': the disconnect between the LLM's textual representation and the separate pixel decoder's latent space, which often leads to inconsistent segmentation masks.

\subsubsection{Qualitative Analysis}
\label{sec:qual}

Illustrations in Figure~\ref{fig:combined_qualitative_results} 
%presents a visual comparison of segmentation predictions
%further illustrate this, and 
show that MAgSeg produces sharp, instance-aware boundaries for agricultural fields where competitors often output noisy or over-smoothed/undersegmented masks.
%Figure~\ref{fig:alu_qualitative_results} presents a visual comparison of segmentation predictions. 
The visual comparisons corroborate our quantitative findings: Text4Seg outputs suffer from severe blockiness due to the aforementioned downsampling artifacts. Similarly, decoder-based methods like LISA often generate artifacts or fail to capture crisp boundaries, likely due to latent misalignment. In contrast, our proposed method generates aesthetically clean, pixel-precise masks that closely align with the ground truth, effectively capturing the nuanced geometries of agricultural fields.
The consistency of these results across both India (ALU) and Southeast Asia (Cambodia and Vietnam) confirms that MAgSeg's superiority is intrinsic to the method, and performs equally well across different landscapes. 
%Qualitative comparisons in 
Figures \ref{fig:alu_qualitative_results}, 
\ref{fig:cambodia_qualitative_results} and \ref{fig:vietnam_qualitative_results} 
in Appendix \ref{sec:more_qual}
show additional illustrations for India, Cambodia and Vietnam.

\subsubsection{Zero Shot Generalization}
\label{sec:zs-results}
To evaluate the generalization capabilities of MAgSeg across different geographical regions, we conduct a zero-shot evaluation on the AI4SmallFarms dataset. Specifically, models trained solely on the ALU dataset (India) were evaluated directly on the Cambodia and Vietnam test sets without any fine-tuning. This setting is particularly challenging due to differences in agricultural landscapes between the source and target domains. The results are presented in Table~\ref{tab:zs-result-ai4sf}. We observe that MAgSeg, the 4B version itself, significantly outperforms decoder-dependent baselines in this regime. For Cambodia, our model achieves a mean IoU of $0.40$, surpassing the strongest baseline, LISA, which achieves $0.25$. Notably, the median IoU for our method is $0.43$, whereas baselines struggle to exceed $0.15$. A similar trend is observed for Vietnam, where our method achieves a mean IoU of $0.41$, compared to $0.36$ for LISAt. These results indicate that decoder-based architectures may overfit to the visual statistics of the training domain, leading to poor performance when subjected to domain shifts. In contrast, our generative approach appears to learn more robust, semantic-aware representations that transfer effectively to unseen agricultural landscapes.
% Please add the following required packages to your document preamble:
% \usepackage{multirow}
% \usepackage{graphicx}
\begin{table}[!h]
\centering
\resizebox{0.8\columnwidth}{!}{%
\begin{tabular}{c|l|rrrrr|r}
\toprule
\textbf{Country} & \multicolumn{1}{c|}{\textbf{Metric}} & \multicolumn{1}{c}{\textbf{LAVT}} & \multicolumn{1}{c}{\textbf{LISA}} & \multicolumn{1}{c}{\textbf{LISAt}} & \multicolumn{1}{c}{\textbf{GSVA}} & \multicolumn{1}{c|}{\textbf{GRES}} & \multicolumn{1}{c}{\textbf{\begin{tabular}[c]{@{}c@{}}MAgSeg (4B) \end{tabular}}} \\ \midrule
\multirow{4}{*}{\textbf{Cambodia}} & \textbf{mean IoU} & 0.09 & \underline{0.25} & 0.22 & 0.21 & 0.20 & \textbf{0.40} \\
 & \textbf{median IoU} & 0.03 & \underline{0.15} & 0.13 & 0.10 & 0.10 & \textbf{0.43} \\
 & \textbf{FNR} & \underline{17.87} & 23.17 & 22.63 & 26.73 & 25.02 & \textbf{15.69} \\
 & \textbf{FPR} & \underline{13.36} & 23.13 & 20.10 & 20.16 & 21.11 & \textbf{11.02} \\ \midrule
\multirow{4}{*}{\textbf{Vietnam}} & \textbf{mean IoU} & 0.13 & 0.30 & \underline{0.36} & 0.29 & 0.33 & \textbf{0.41} \\
 & \textbf{median IoU} & 0.05 & 0.19 & \underline{0.30} & 0.19 & 0.22 & \textbf{0.37} \\
 & \textbf{FNR} & 19.98 & 24.10 & \textbf{9.05} & 26.98 & 12.57 & \underline{10.33} \\
 & \textbf{FPR} & \underline{30.02} & 50.48 & 48.41 & 46.42 & 47.19 & \textbf{25.63} \\ \bottomrule
\end{tabular}%
}
\caption{Zero-shot evaluation on field boundary segmentation: trained on India and evaluated directly on Cambodia and Vietnam.}
\label{tab:zs-result-ai4sf}
\end{table}

\subsubsection{Architectural and Training Efficiency}
\label{sec:efficiency}

A critical advantage of our approach is highlighted by the \textit{Overhead} metric (Table~\ref{tab:main-result}, top row). We observe a clear dichotomy in prior work: decoder-free methods like Text4Seg offer low overhead but poor performance, whereas decoder-based methods achieve better performance at the cost of high overhead. Our method successfully breaks this trade-off, delivering superior segmentation performance with zero decoder overhead. This confirms that proper instruction tuning (via LoRA) and reinforcement learning (via GRPO) are sufficient to ground LLMs in pixel-level tasks without parameter-heavy auxiliary networks.

We also analyze the %architectural and 
computational efficiency of MAgSeg with the best-performing baselines in Table~\ref{tab:avg-train-time-per-token}. 
To ensure a fair comparison across different base model sizes, we report a \textit{parameter-normalized training time}, defined as the average training duration per token divided by the total number of model parameters. Our decoder-free approach demonstrates superior efficiency on both metrics. First, regarding architectural complexity, our method incurs zero overhead, whereas decoder-dependent baselines introduce substantial additional parameters, ranging from $9.35\%$ (GSVA) to $15.82\%$ (GRES). Second, this streamlined architecture translates directly to training speed. Our method achieves a normalized training time of $1.33 \times 10^{-8}$ s, significantly outperforming LISA/LISAt ($1.69$), GRES ($1.80$), and GSVA ($2.04$).
This reduction in latency confirms that eliminating auxiliary vision decoders not only simplifies deployment but also enhances the computational throughput of the training process.

% Please add the following required packages to your document preamble:
% \usepackage{graphicx}
\begin{table}[!h]
\centering
\resizebox{0.8\columnwidth}{!}{%
\begin{tabular}{c|ccc|c}
\toprule
 & \textbf{LISA} & \textbf{GSVA} & \textbf{GRES} & \textbf{MAgSeg} \\ \midrule
\textbf{Overhead} & $9.48\%$ & $\underline{9.35}\%$ & $15.82\%$ & $\textbf{0.00}\%$  \\ 
\begin{tabular}[c]{@{}c@{}}\textbf{Average Time} (s $\times 10^{-8})$\end{tabular} & \underline{1.69} & 2.04 & 1.80 & \textbf{1.33} \\ \toprule
\end{tabular}%
}
\caption{Computational efficiency: parameter overhead ($\downarrow$) and the parameter-normalized training time per token ($\downarrow$) of MAgSeg and the best baseline methods. Other overhead values are in Table \ref{tab:main-result}.}
\label{tab:avg-train-time-per-token}
\end{table}

\subsubsection{Additional Results}
\label{sec:additional_results}
We evaluate the impact of GRPO post training and EPOC refinement during inference in MAgSeg through an \textbf{ablation study}.
Table \ref{tab:alu-ablation} in Appendix \ref{sec:ablation}
shows that, while each of them individually improves the mean IoU, the combination of both components yields the best performance.
We qualitatively demonstrate the importance of GRPO in suppressing spurious boundaries in figure \ref{fig:grpo_qual_comp} (Appendix \ref{sec:grpo_visualization}).
We also present (in Appendix \ref{sec:stratified_results}) the comparative performance on field segmentation \textbf{stratified by various criteria}: MAgSeg outperforms the best baselines across 5 different \textit{field size ranges} ($<100 m^2$ to 5 acres; and by a large margin for smaller field sizes $< 1$ acre), 5 \textit{climatic regions} and 7 \textit{ecological regions}.
MAgSeg's robustness to field sizes is particularly important in  smallholder landscapes, where plot sizes are highly heterogeneous.

\subsubsection{Limitations} 
\label{sec:limitation} 
Our experiments, which demonstrate MAgSeg's state-of-the-art performance, also highlight its limitations.
%We observe a performance disparity between majority and minority classes. 
While it excels at segmenting majority classes (\textit{fields}), MAgSeg exhibits reduced efficacy on minority classes (e.g., \textit{wells, ponds}), a challenge common to many segmentation baselines due to class imbalance inherent in the dataset itself. 
Additionally, our independent patch-based inference strategy prevents MAgSeg from leveraging global spatial context \textit{across} patch boundaries. This lack of cross-patch information can lead to stitching artifacts and geometric discontinuities in the final maps (see figure \ref{fig:error_analysis} in Appendix \ref{sec:vis_limitations}). Future work can extend our framework to address these limitations.

\section{Conclusion}
\label{sec:conclusion}
In this work, we presented MAgSeg, the first MLLM framework specifically adapted for smallholder agricultural landscape segmentation from VHR satellite imagery. 
%By casting segmentation as a pure text-generation task, 
MAgSeg successfully addresses the critical bottlenecks of prior decoder-free MLLM approaches -- token-length limitations and domain alignment inefficiencies,  
% Our two-stage training paradigm combines patch-aware LoRA fine tuning with reward-driven GRPO post-training, on a novel visual instruction data format, 
% enabling standard LLMs to achieve pixel-perfect segmentation 
without using parameter-heavy auxiliary decoders. 
Extensive evaluations on smallholder agriculture datasets spanning 3 countries demonstrate the superiority of our method %establishes a new state-of-the-art among 
over previous MLLM-based methods, achieving up to %$21$-point mIoU 
$4\times$ 
improvement over the nearest competitor %on ALU dataset 
while incurring zero parameter overhead. 
Our evaluation
%on the AI4SmallFarms dataset
also confirms the robustness of MAgSeg across diverse geographies including in zero-shot settings.
%, where it outperformed baselines by nearly $4\times$ in challenging terrains. 
Ultimately, this work illustrates that with appropriate grounding strategies, general-purpose foundation models can be effectively adapted for specialized, high-precision agricultural landscape segmentation, paving the way for scalable, accessible, and accurate global agricultural monitoring.

%\newpage
% \newpage
%% The file named.bst is a bibliography style file for BibTeX 0.99c
\bibliographystyle{named}
\bibliography{ijcai25,ref}

\newpage

\appendix

\section{RRLE Encoding and Decoding}
\label{sec:rrle_algorithm}
We present the pseudo-code for converting a semantic mask into its purely text-based RRLE representation in Algorithm~\ref{alg:rle_encoding}. The encoder processes each row of the mask and expresses it in the format `value * count', where \textit{value} denotes the class label and \textit{count} denotes the length of the corresponding contiguous run of pixels. This transformation is fully invertible: Algorithm~\ref{alg:rle_decoding} provides the corresponding decoder that reconstructs the semantic mask from a given RRLE string. This bidirectional mapping is crucial in Stage~2 of our framework, where we convert the model’s generated response back into a semantic mask to compute the DICE score for the GRPO objective.

\begin{algorithm}[t]
\caption{Row Run-Length Encoding (RRLE)}
\label{alg:rle_encoding}
\begin{algorithmic}[1]
    \Require Semantic Mask, $\rvr \in \mathbb{R}^{h_\rvp \times w_\rvp}$
    \Ensure RRLE string, $\rvt$
    
    \Procedure{\texttt{rrle}}{$\rvr$}
        \State $\rvt_{rows} \leftarrow \emptyset$ 
        \State \textit{$\triangleright$ Iterate through rows}
        \For{$i \leftarrow 0$ \textbf{to} $h_\rvp - 1$}
            \State $R \leftarrow \rvr[i]$
            \State $\rvt_{parts} \leftarrow \emptyset, \quad v \leftarrow R[0], \quad c \leftarrow 1$
            
            \State \textit{$\triangleright$ Iterate through columns}
            \For{$j \leftarrow 1$ \textbf{to} $w_\rvp - 1$}
                \If{$R[j] = v$}
                    \State $c \leftarrow c + 1$
                \Else
                    \State \textit{$\triangleright$ Format: `value * count'}
                    \State $\rvt_{parts} \leftarrow \rvt_{parts} \cup \{ v ~~ \text{*} c \}$
                    \State $v \leftarrow R[j], \quad c \leftarrow 1$
                \EndIf
            \EndFor
            
            \State \textit{$\triangleright$ Handle the final run of the row}
            \State $\rvt_{parts} \leftarrow \rvt_{parts} \cup \{ v ~~ \text{*} c \}$
            
            \State $\rvt_{row} \leftarrow \text{join}(\rvt_{parts}, \text{`}|\text{'})$
            \State $\rvt_{rows} \leftarrow \rvt_{rows} \cup \{ \rvt_{row} \}$
        \EndFor
        
        \State \Return $\rvt = \text{join}(\rvt_{rows}, \text{`\textbackslash n'})$
    \EndProcedure
\end{algorithmic}
\end{algorithm}

\begin{algorithm}[t]
\caption{Decode RRLE String to Mask}
\label{alg:rle_decoding}
\begin{algorithmic}[1]
    \Require RRLE String $\rvt$, Class Map $\mathcal{M}$, Dimensions $h_\rvp, w_\rvp$
    \Ensure Reconstructed Semantic Mask, $\rvr \in \mathbb{R}^{h_\rvp \times w_\rvp}$
    
    \Procedure{\texttt{decode\_rrle}}{$\rvt, \mathcal{M}, h_\rvp, w_\rvp$}
        \State $\rvr_{flat} \leftarrow \mathbf{0}_{h_\rvp \cdot w_\rvp}$ \Comment{Initialize flat zero array}
        \State $p \leftarrow 0$ \Comment{Current pixel pointer}
        
        \State $\rvt_{rows} \leftarrow \text{split}(\rvt, \text{`\textbackslash n'})$
        
        \State \textit{$\triangleright$ Process each row from the encoded string}
        \For{\textbf{each} $s_{row} \in \rvt_{rows}$}
            \If{$s_{row}$ is empty}
                \State $p \leftarrow p + w_\rvp$ \Comment{Skip empty row}
                \State \textbf{continue}
            \EndIf
            
            \State $\rvt_{tokens} \leftarrow \text{split}(s_{row}, \text{`}|\text{'})$
            
            \State \textit{$\triangleright$ Decode runs within the row}
            \For{\textbf{each} $t \in \rvt_{tokens}$}
                \State $(l, c_{str}) \leftarrow \text{Split}(t, \text{`*'}) $
                \State $c \leftarrow \text{integer}(c_{str})$
                \State $v \leftarrow \mathcal{M}.\text{get}(l, 0)$ \Comment{Map label string to ID}
                
                \State \textit{$\triangleright$ Safety bound check}
                \If{$p + c > h_\rvp \cdot w_\rvp$}
                    \State $c \leftarrow (h_\rvp \cdot w_\rvp) - p$
                \EndIf
                
                \State $\rvr_{flat}[p : p + c] \leftarrow v$ \Comment{Fill run interval}
                \State $p \leftarrow p + c$
            \EndFor
        \EndFor
        
        \State \Return $\rvr = \text{reshape}(\rvr_{flat}, (h_\rvp, w_\rvp))$
    \EndProcedure
\end{algorithmic}
\end{algorithm}

\section{Implementation Details}
\label{sec:implementation}
\noindent\textbf{MAgSeg.} 
%For our method, we consider Gemma3 series~\cite{gemma} of models. Particularly, we consider two sizes of Gemma - Gemma3-4B and Gemma3-12B. 
In the initial SFT stage, we finetune the model for $20$ epochs with usual next-token prediction cross-entropy loss. The LoRA rank is set to $8$, alpha is set to $16$. The learning rate is set to $2\times10^{-4}$ with a batch size of $8$ with $4$ gradient accumulation steps. We use $h_\rvp = w_\rvp = 32$ for patch dimension as mentioned in Section~\ref{sec:methodology}. 
Further, in the GRPO Post-training, we finetune the model using the GRPO loss as mentioned in Eq.~\ref{eq:pt_loss} with $\epsilon=10^{-3}$ and $\beta=10^{-4}$. We use learning rate of $10^{-6}$ with cosine-schedule and weight decay of $0.1$.   Further, for each sample, we generate $G=24$ responses (rollouts) which are then assigned a reward as illustrated in Eq.~\ref{eq:dice_reward} to calculate the final loss objective. 

%\noindent\textbf{Baseline Details.} 

\noindent\textbf{LAVT.} LAVT uses a Swin transformer vision backbone and general BERT transformer backbone. It employs AdamW optimizer with weight decay of $0.01$, learning rate of $5\times10^{-5}$ for the Swin transformer backbone and the BERT backbone. Further, a polynomial decay with power of $0.9$ is used as LR scheduler, batch size is set to $8$, the gradient clipping is also enabled with a threshold of $0.01$. The model is trained for a total of $20$ epochs.

\noindent\textbf{LISA.} LISA uses a 13B parameter base LLM derived from LLaVA architecture (specifically LLaVA-Llama-2-13B), CLIP ViT-L/14 vision tower, and SAM decoder head with ViT-Huge backbone. The learning rate is set to $3\times10^{-4}$ with AdamW optimizer with WarmupDecayLR scheduler with $100$ warmup steps. The LoRA rank is set to $8$, alpha is set to $16$, the dropout rate is set to $5\%$. The cross-entropy loss, BCE mask loss and DICE mask loss are assigned weights of $1.0$, $2.0$ and $0.5$ respectively. The model is trained for total of $10$ epochs with batch size of $2$.

\noindent\textbf{LISAt.} LISAt uses a 7B parameter base LLM derived from LLaVa architecture, CLIP ViT-L/14 vision tower, and SAM decoder head with ViT-Huge backbone. The learning rate is set to $3\times10^{-4}$ with AdamW optimizer with WarmupDecayLR scheduler with $100$ warmup steps. The LoRA rank is set to $8$, alpha is set to $16$, the dropout rate is set to $5\%$. The cross-entropy loss, BCE mask loss and DICE mask loss are assigned weights of $1.0$, $2.0$ and $0.5$ respectively. The model is trained for total of $60$ epochs with batch size of $12$.

\noindent\textbf{GSVA.} GSVA uses a 13B parameter base LLM (by default) derived from LLaVA architecture, CLIP ViT-L/14 vision tower, and SAM decoder head with ViT-Huge backbone. A key addition is the [REJ] token to explicitly reject null targets. The learning rate is set to $3\times10^{-4}$ with AdamW optimizer with WarmupDecayLR scheduler with $100$ warmup steps. The LoRA rank defaults to $8$ in code, with alpha set to $16$ and dropout rate set to $5\%$. The cross-entropy loss, BCE mask loss and DICE mask loss are assigned weights of $1.0$, $2.0$ and $0.5$ respectively. The model is trained for total of $10$ epochs with batch size of $20$.

\noindent\textbf{GRES.} GRES uses a Swin-Base transformer as the vision backbone and a BERT-Base transformer as the linguistic backbone to explicitly model region-region and region-language dependencies. The model is trained using the AdamW optimizer with a weight decay of $0.05$ and a learning rate of $1\times10^{-5}$. The model is trained for $~600$ epochs with a batch size of $8$.

\noindent\textbf{Text4Seg.} Text4Seg uses Vicuna-7B as the base LLM and CLIP ViT-L/14 vision tower. Since it is a decoder-free method, it doesn't contain any decoder head. The learning rate is set to $10^{-5}$ with cosine-schedule and weight decay of $0.01$. The model is trained for $50$ epochs with batch size of $8$. The refined predictions are obtained by refining the coarse predictions using SAM ViT-H backbone.

\section{Evaluation Metrics}
\label{sec:metrics_defn}

Following \cite{alu}, for instance-wise metrics, we merge and match multiple predictions that overlap with a ground truth instance. A particular predicted instance is considered to match with a ground truth instance if they belong to the same class and overlap with the ground truth instance. Then the per-instance IoU is defined as:
\begin{align}
    \text{IoU}_I = \frac{P^i_m \cap P^i}{P^i_m \cup P^i}
\label{eq:iou_metric}
\end{align}
where, $P^i_m$ is the total number of pixels in all the predicted i$^{th}$ merged instance and $P^i$ is the the total number of pixels in the ground truth i$^{th}$ instance. We compute the mean/median IoU by averaging (or taking the median of) the per-instance IoU scores across all instances in a given image $I$. This instance-level IoU measures the degree of overlap between predicted and ground-truth masks, where higher mIoU indicates stronger segmentation performance.We compute the mean/median IoU by averaging (or taking the median of) the per-instance IoU scores across all instances in a given image $I$. This instance-level IoU measures the degree of overlap between predicted and ground-truth masks, where higher mIoU indicates stronger segmentation performance.

Further, FNR is defined as the fraction of ground truth instances that do not have any matching predicted instance. Similarly, FPR is defined as the fraction of predicted instances that are not matches with a ground truth instance.

% Further, we also use PoLiS score~\cite{PERSELLO2019111253} to assess the quality of field delienation. 
% Particularly, if $A$ and $B$ are two polygons representing the ground truth and the predicted field, respectively. Let $\partial A$ denote the boundary of polygon $A$, represented by an ordered set of $N_A$ vertices $\{a_1, a_2, \dots, a_{N_A}\}$. Similarly, let $\partial B$ be the boundary of polygon $B$ with $N_B$ vertices $\{b_1, b_2, \dots, b_{N_B}\}$. The PoLiS score $d_{\text{PoLiS}}(A, B)$ is defined as the symmetric average distance between the vertices of one polygon and the boundary of the other. It is calculated as the sum of two directional distances:
% \begin{align}
%     d_{\text{PoLiS}}(A, B) &= \frac{1}{2 N_A} \sum_{i=1}^{N_A} \min_{q \in \partial B} \| a_i - q \| \nonumber\\
%     &~~~~~~~+ \frac{1}{2 N_B} \sum_{j=1}^{N_B} \min_{p \in \partial A} \| b_j - p \| 
% \label{eq:polis_score}
% \end{align}
% where, $\min_{q \in \partial B} \| a_i - q \|$ is the shortest distance from the vertex $a_i$ to any point $q$ lying on the boundary segments of $B$. The PoLiS score represents the average displacement between the boundaries. A lower score indicates higher geometric similarity, with a score of $0$ representing a perfect match.

\section{Additional Results}
\label{sec:more_results}

\subsection{Ablation Studies}
\label{sec:ablation}
To isolate the contributions of specific components in our framework, we conduct an ablation study on the ALU dataset, evaluating the impact of the EPOC refinement process during inference and the GRPO post-training stage (Table~\ref{tab:alu-ablation}). 
EPOC primarily targets refinements of boundaries of the agricultural fields; applying it to the SFT baseline yields a substantial improvement in the \textit{Fields} class (e.g., mean IoU increases from $0.35$ to $0.47$ for MAgSeg-4B). 
However, EPOC alone provides negligible gains for other classes such as \textit{Trees}, \textit{Ponds}, and \textit{Wells}. This is because while boundaries of fields are prominent, boundaries of other classes are not so prominent and hence it does not help with the refinement.  
Further, the inclusion of GRPO introduces a significant performance uplift across multiple categories, notably boosting \textit{Trees} from $0.08$ to $0.14$ and enabling the detection of \textit{Ponds} ($0.05$). This indicates that the RL objective, by optimizing for the pixel-level DICE score, successfully grounds the model in the visual semantics of the satellite imagery, capturing finer details that supervised tuning misses. We provide qualitative results illustrating the effect of GRPO phase in Figure~\ref{fig:grpo_qual_comp} %in Appendix. 
The combination of both components yields the highest performance, demonstrating their complementary nature. 
For the MAgSeg-12B model, the full method achieves a mean IoU of $0.59$ for \textit{Fields} and $0.21$ for \textit{Trees}. While GRPO enhances the model's fundamental ability to localize and classify features, EPOC refines the boundaries of the generated masks, resulting in superior overall segmentation. Similar trends are observed across both the 4B and 12B model scales, confirming the robustness of the proposed pipeline.
\begin{table}[!ht]
\centering
\renewcommand{\arraystretch}{1.5} 
\resizebox{\columnwidth}{!}{%
\begin{tabular}{ccclccccc}
\toprule
 & \multicolumn{2}{c}{\textbf{Components}} & & \multicolumn{5}{c}{\textbf{mean IoU}} \\ 
\cmidrule{2-3} \cmidrule{5-9} 
\textbf{Model} & \textbf{EPOC} & \textbf{GRPO} & & \textbf{Fields} & \textbf{Trees} & \textbf{Clouds} & \textbf{Ponds} & \textbf{Wells} \\ \midrule
% Block 1: Gemma3-4B
\multirow{4}{*}{\rotatebox[origin=c]{90}{\textbf{MAgSeg-4B}}} & {\color{red}\xmark} & {\color{red}\xmark} & & 0.35 & 0.08 & 0.20 & 0.00 & 0.00 \\
 & {\color{green}\cmark} & {\color{red}\xmark} & & 0.47 & 0.08 & 0.21 & 0.00 & 0.00 \\
 & {\color{red}\xmark} & {\color{green}\cmark} & & 0.46 & 0.14 & 0.23 & 0.05 & 0.00 \\
 & {\color{green}\cmark} & {\color{green}\cmark} & & 0.58 & 0.20 & 0.25 & 0.05 & 0.00 \\ \midrule
% Block 2: Gemma3-12B
\multirow{4}{*}{\rotatebox[origin=c]{90}{\textbf{MAgSeg-12B}}} & {\color{red}\xmark} & {\color{red}\xmark} & & 0.39 & 0.11 & 0.20 & 0.00 & 0.00 \\
 & {\color{green}\cmark} & {\color{red}\xmark} & & 0.50 & 0.11 & 0.20 & 0.00 & 0.00 \\
 & {\color{red}\xmark} & {\color{green}\cmark} & & 0.54 & 0.14 & 0.22 & 0.05 & 0.00 \\
 & {\color{green}\cmark} & {\color{green}\cmark} & & 0.59 & 0.21 & 0.25 & 0.05 & 0.00 \\ \bottomrule
\end{tabular}%
}
\caption{Ablation study on the ALU dataset to analyze the impact of EPOC refinement and GRPO post-training on instance-wise mean IoU ($\uparrow$) across all classes, for both 4B and 12B models.}
\label{tab:alu-ablation}
\end{table}

\subsection{Stratified Results}
\label{sec:stratified_results}
To assess the model's efficacy under various agricultural conditions, we present stratified results under three stratification criteria - (i) agricultural field size, (ii) climatic region, and (iii) ecological region. 

\noindent\textit{(i) Field Size.} We present stratified results across varying agricultural field sizes, we present a size-stratified performance analysis in Figure~\ref{fig:alu_size_stratified_results}. We observe performance gain for our proposed method on smaller field segments ($< 2$ acres), where baseline methods often struggle to resolve fine-grained boundaries. This margin is most significant in the critical $< 500 \, \text{m}^2$ and $500 \, \text{m}^2 - 1 \, \text{acre}$ brackets, which constitute the majority of smallholder plots. As field size increases ($> 5$ acres), our method maintains robust performance, achieving parity with the strongest baselines. The stability of our results across the intermediate range ($500 \, \text{m}^2$ to $5$ acres) underscores the method's robustness to field sizes, confirming its suitability for diverse smallholder environments where plot sizes are highly heterogeneous.

\begin{figure}[!ht] % Use figure* for full-width in two-column papers
    \centering
    % Adjust the width if you want margins (e.g., 0.95\linewidth)
    \resizebox{\columnwidth}{!}{ 
        % Add these to your preamble if not already defined

\centering
\begin{tikzpicture}
    \begin{axis}[
        ybar,
        width=\textwidth,
        height=7.5cm,
        bar width=11pt, 
        ymin=0, ymax=0.8,
        ylabel={\textbf{mean IoU}},
        xtick={0,1,2,3,4,5},
        xticklabels={
            {\large \textbf{I} \\ (4,909)}, 
            {\large \textbf{II} \\ (271)}, 
            {\large \textbf{III} \\ (470)}, 
            {\large \textbf{IV} \\ (10,582)}, 
            {\large \textbf{V} \\ (7,399)}, 
            {\large \textbf{VI} \\ (40)}
        },
        x tick label style={
            font=\small\bfseries, 
            text width=2.0cm,   
            align=center,
            yshift=-4pt         
        },
        ylabel style={font=\bfseries},
        ymajorgrids=true,
        grid style={dashed, gray!40},
        enlarge x limits=0.12, 
        legend style={at={(0.5,1.15)}, anchor=south, legend columns=-1, draw=none, fill=none, /tikz/every even column/.append style={column sep=0.3cm}}
    ]

    % --- Method: Ours (Gemma4Seg_GRPO) ---
    \addplot[fill=oursred, draw=oursred!50!black] coordinates {
        (0, 0.444) (1, 0.623) (2, 0.203) (3, 0.516) (4, 0.449) (5, 0.074)
    };

    % --- Method: GRES ---
    \addplot[
        pattern=north east lines, 
        pattern color=gresblue!60, 
        draw=gresblue,
        preaction={fill=gresblue!20}
    ] coordinates {
        (0, 0.299) (1, 0.454) (2, 0.068) (3, 0.369) (4, 0.354) (5, 0.094)
    };

    % --- Method: LiSAT ---
    \addplot[
        pattern=crosshatch, 
        pattern color=lisatorange!70,
        draw=lisatorange,
        preaction={fill=lisatorange!20}
    ] coordinates {
        (0, 0.275) (1, 0.467) (2, 0.073) (3, 0.343) (4, 0.329) (5, 0.076)
    };

    % --- Method: GSVA ---
    \addplot[
        pattern=dots, 
        pattern color=gsvagreen!70,
        draw=gsvagreen,
        preaction={fill=gsvagreen!20}
    ] coordinates {
        (0, 0.265) (1, 0.515) (2, 0.069) (3, 0.353) (4, 0.314) (5, 0.035)
    };

    % --- Method: Text4Seg (R) ---
    \addplot[
        pattern=horizontal lines, 
        pattern color=textviolet!70,
        draw=textviolet,
        preaction={fill=textviolet!20}
    ] coordinates {
        (0, 0.059) (1, 0.098) (2, 0.025) (3, 0.072) (4, 0.070) (5, 0.018)
    };

    \legend{\textbf{Ours}, \textbf{GRES}, \textbf{LISAt}, \textbf{GSVA}, \textbf{Text4Seg (R)}}
    \end{axis}
\end{tikzpicture}
    }
    \caption{Climatic Region stratified performance analysis on the ALU dataset. Labels \textbf{I--VI} correspond to: 
    \textbf{I}: Central Plateau Region; 
    \textbf{II}: Lower Gangetic Region; 
    \textbf{III}: Southern Plateau Region; 
    \textbf{IV}: Trans Gangetic Region; 
    \textbf{V}: Upper Gangetic Region; 
    \textbf{VI}: Western Himalayan Region; 
    Total instance counts ($N$) for each region are provided in parentheses on the x-axis.}
    \label{fig:alu_climate_stratified_results}
\end{figure}
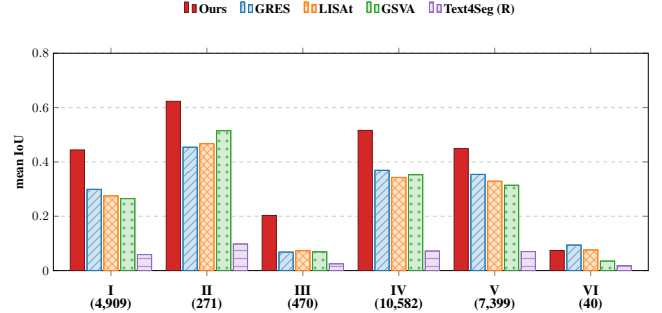

\noindent\textit{(ii) Climatic Region.} Climatic conditions play a crucial role in influencing the geometry of agricultural fields. For instance, monsoonal floodplains like the Upper and Lower Gangetic regions receive heavy rainfall with gentle slopes, fostering regular rectilinear fields, whereas high-altitude zones like the Western Himalayas necessitate terrace-based, narrow, and elongated fields. To evaluate the robustness of different methods under these topological variations, we present a climatic region-stratified comparison in Figure~\ref{fig:alu_climate_stratified_results}. Our proposed method demonstrates consistent superiority across the major climatic zones (I--V). Notably, in the most data-abundant regions (IV and V), our model maintains a substantial lead, improving mean IoU by approximately $0.15$ over the best performing baselines (GRES and LISAt). This advantage is even more pronounced in the challenging Region III, where our method ($0.21$) nearly triples the performance of the next best competitor ($0.07$), highlighting its capability to resolve complex field boundaries in Southern Plateau regions where baseline methods falter. While Region II yields the highest absolute scores across all methods, our approach effectively maximizes this advantage, achieving mean IoU of $0.623$. The stability of these results validates the methods's generalization capability across diverse agro-climatic typologies.

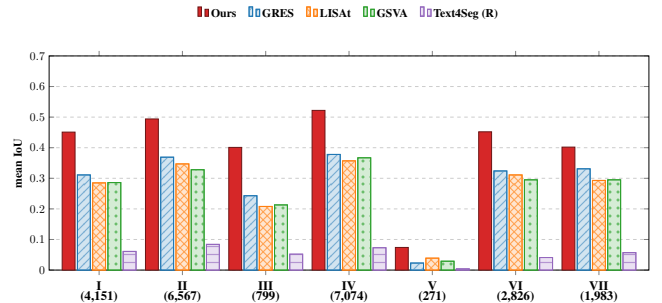
\begin{figure}[!ht] % Use figure* for full-width in two-column papers
    \centering
    % Adjust the width if you want margins (e.g., 0.95\linewidth)
    \resizebox{\columnwidth}{!}{ 
        \centering
\begin{tikzpicture}
    \begin{axis}[
        ybar,
        width=1.1\textwidth, % Slightly wider to accommodate more categories
        height=8cm,
        bar width=11pt, % Thinner bars for 7 categories
        ymin=0, ymax=0.7,
        ylabel={\textbf{mean IoU}},
        xtick={0,1,2,3,4,5,6},
        xticklabels={
            {\large \textbf{I} \\ (4,151)}, 
            {\large \textbf{II} \\ (6,567)}, 
            {\large \textbf{III} \\ (799)}, 
            {\large \textbf{IV} \\ (7,074)}, 
            {\large \textbf{V} \\ (271)}, 
            {\large \textbf{VI} \\ (2,826)}, 
            {\large \textbf{VII} \\ (1,983)}
        },
        x tick label style={
            font=\tiny\bfseries, % Smaller font for long full names
            text width=2.2cm,   
            align=center,
            yshift=-2pt         
        },
        ylabel style={font=\bfseries},
        ymajorgrids=true,
        grid style={dashed, gray!40},
        enlarge x limits=0.1, 
        legend style={at={(0.5,1.15)}, anchor=south, legend columns=-1, draw=none, fill=none, /tikz/every even column/.append style={column sep=0.3cm}}
    ]

    % --- Method: Ours (Gemma4Seg_GRPO) ---
    \addplot[fill=oursred, draw=oursred!50!black] coordinates {
        (0, 0.451) (1, 0.494) (2, 0.401) (3, 0.522) (4, 0.074) (5, 0.452) (6, 0.402)
    };

    % --- Method: GRES ---
    \addplot[
        pattern=north east lines, 
        pattern color=gresblue!60, 
        draw=gresblue,
        preaction={fill=gresblue!20}
    ] coordinates {
        (0, 0.311) (1, 0.369) (2, 0.243) (3, 0.378) (4, 0.023) (5, 0.324) (6, 0.331)
    };

    % --- Method: LiSAT ---
    \addplot[
        pattern=crosshatch, 
        pattern color=lisatorange!70,
        draw=lisatorange,
        preaction={fill=lisatorange!20}
    ] coordinates {
        (0, 0.285) (1, 0.347) (2, 0.208) (3, 0.357) (4, 0.039) (5, 0.311) (6, 0.293)
    };

    % --- Method: GSVA ---
    \addplot[
        pattern=dots, 
        pattern color=gsvagreen!70,
        draw=gsvagreen,
        preaction={fill=gsvagreen!20}
    ] coordinates {
        (0, 0.286) (1, 0.328) (2, 0.213) (3, 0.367) (4, 0.029) (5, 0.295) (6, 0.295)
    };

    % --- Method: Text4Seg (R) ---
    \addplot[
        pattern=horizontal lines, 
        pattern color=textviolet!70,
        draw=textviolet,
        preaction={fill=textviolet!20}
    ] coordinates {
        (0, 0.061) (1, 0.084) (2, 0.052) (3, 0.073) (4, 0.004) (5, 0.041) (6, 0.057)
    };

    \legend{\textbf{Ours}, \textbf{GRES}, \textbf{LISAt}, \textbf{GSVA}, \textbf{Text4Seg (R)}}
    \end{axis}
\end{tikzpicture}
    }
    \caption{Ecological Region stratified performance analysis on the ALU dataset. Labels \textbf{I--VII} correspond to: 
    \textbf{I}: Assam and Bengal Plain Hot Subhumid to Humid Eco-Region; 
    \textbf{II}: Central Highlands (Malwa), Gujarat Plain, and Kathiawar Peninsula Semi-Arid Eco-Region; 
    \textbf{III}: Central Highlands (Malwa and Bundelkhand) Hot Subhumid (Dry) Eco-Region; 
    \textbf{IV}: Deccan Plateau Hot Semi-Arid Eco-Region; 
    \textbf{V}: North Eastern Hills (Purvachal) Warm Perhumid Eco-Region; 
    \textbf{VI}: Northern Plain (and Central Highlands) Including Aravallis Hot Semi-Arid Eco-Region; 
    \textbf{VII}: Western Plain Kachchh and part of Kathiawar Peninsula Hot Arid Eco-Region. 
    Total instance counts ($N$) for each region are provided in parentheses on the x-axis.}
    \label{fig:alu_eco_stratified_results}
\end{figure}

\noindent\textit{(iii) Ecological Region.} Agricultural landscapes vary significantly across ecological zones, where factors such as vegetation density, soil composition, and crop distinctiveness alter the visual characteristics of field boundaries. To evaluate generalization across these biomes, we present performance stratified by ecological region in Figure~\ref{fig:alu_eco_stratified_results}. Our method demonstrates remarkable stability, outperforming baselines across all seven identified zones. In the most distinct and data-rich category (Region IV), our model achieves a mean IoU of $0.522$, surpassing the nearest competitor (GRES) by a substantial margin of $\sim0.14$. Furthermore, in Region III, where baseline methods suffer significant degradation (dropping to $\sim0.20-0.24$), our approach maintains a robust performance of $0.40$, nearly doubling the accuracy of LISAt. This consistent superiority across both dominant (Regions I, II, VI, VII) and minority ecological zones confirms that our feature representation remains discriminative regardless of the underlying biological landscape.

\begin{figure}[!ht] % Use figure* for full-width in two-column papers
    \centering
    % Adjust the width if you want margins (e.g., 0.95\linewidth)
    \resizebox{\columnwidth}{!}{ 
        \centering
\begin{tikzpicture}
    \begin{axis}[
        ybar,
        width=\textwidth,
        height=6.5cm,
        % KEY CHANGE: Reduced bar width from 12pt to 8pt
        % This shrinks the clusters, creating more white space between groups
        bar width=11pt, 
        ymin=0, ymax=0.8,
        ylabel={\textbf{mean IoU}},
        xtick={0,1,2,3,4,5},
        xticklabels={
            {$\bf{<100\text{m}^2}$ \\ ~~$(166)$}, 
            {$\bf{100\text{-}500\text{m}^2}$ \\ ~~$(3,567)$}, 
            {$\bf{500\text{m}^2\text{-}1\text{acre}}$ \\ ~~~$(15,496)$}, 
            {$\bf{1\text{-}2\text{acre}}$ \\ $(3,570)$}, 
            {$\bf{2\text{-}5\text{acre}}$ \\ $(1,891)$}, 
            {$\bf{>5\text{acre}}$ \\ ~~$(275)$}
        },
        % Label styling for clean gaps
        x tick label style={
            font=\normalsize\bfseries, 
            text width=1.3cm,   
            align=center,
            yshift=-4pt         
        },
        ylabel style={font=\bfseries},
        ymajorgrids=true,
        grid style={dashed, gray!40},
        % Ensures the plot doesn't feel cramped at the far ends
        enlarge x limits=0.15, 
        legend style={at={(0.5,1.15)}, anchor=south, legend columns=-1, draw=none, fill=none, /tikz/every even column/.append style={column sep=0.5cm}}
    ]

    % --- Method: Ours (Solid Red) ---
    \addplot[fill=oursred, draw=oursred!50!black] coordinates {
        (0, 0.1188) (1, 0.3496) (2, 0.5608) (3, 0.6598) (4, 0.6692) (5, 0.6342)
    };

    % --- Method: GRES (Blue with Pattern) ---
    \addplot[
        pattern=north east lines, 
        pattern color=gresblue!60, 
        draw=gresblue,
        preaction={fill=gresblue!20}
    ] coordinates {
        (0, 0.0187) (1, 0.1317) (2, 0.3495) (3, 0.5497) (4, 0.6293) (5, 0.6886)
    };

    % --- Method: LiSAT (Orange with Pattern) ---
    \addplot[
        pattern=crosshatch, 
        pattern color=lisatorange!70,
        draw=lisatorange,
        preaction={fill=lisatorange!20}
    ] coordinates {
        (0, 0.0223) (1, 0.1335) (2, 0.3429) (3, 0.5083) (4, 0.5787) (5, 0.6145)
    };

    % --- Method: GSVA (Green with Pattern) ---
    \addplot[
        pattern=dots, 
        pattern color=gsvagreen!70,
        draw=gsvagreen,
        preaction={fill=gsvagreen!20}
    ] coordinates {
        (0, 0.0157) (1, 0.1247) (2, 0.3399) (3, 0.5176) (4, 0.5646) (5, 0.6303)
    };

    % --- Method: Text4Seg (Violet with Pattern) ---
    \addplot[
        pattern=horizontal lines, 
        pattern color=textviolet!70,
        draw=textviolet,
        preaction={fill=textviolet!20}
    ] coordinates {
        (0, 0.0053) (1, 0.0275) (2, 0.0659) (3, 0.1116) (4, 0.1480) (5, 0.2132)
    };

    \legend{\textbf{Ours}, \textbf{GRES}, \textbf{LISAt}, \textbf{GSVA}, \textbf{Text4Seg (R)}}
    \end{axis}
\end{tikzpicture}
    }
    \caption{Size-stratified performance analysis on the ALU dataset. Total instance counts ($N$) for each region are provided in parentheses on the x-axis.}
    \label{fig:alu_size_stratified_results}
\end{figure}
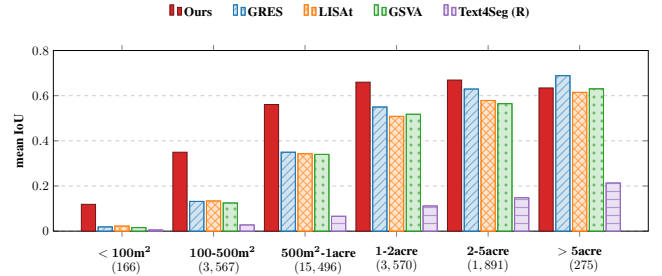

\subsection{Qualitative Analysis: Effect of GRPO}
\label{sec:grpo_visualization}

\begin{figure*}[t] % Use figure* for full-width in two-column papers
    \centering
    % Adjust the width if you want margins (e.g., 0.95\linewidth)
    \resizebox{\textwidth}{!}{ 
        \input{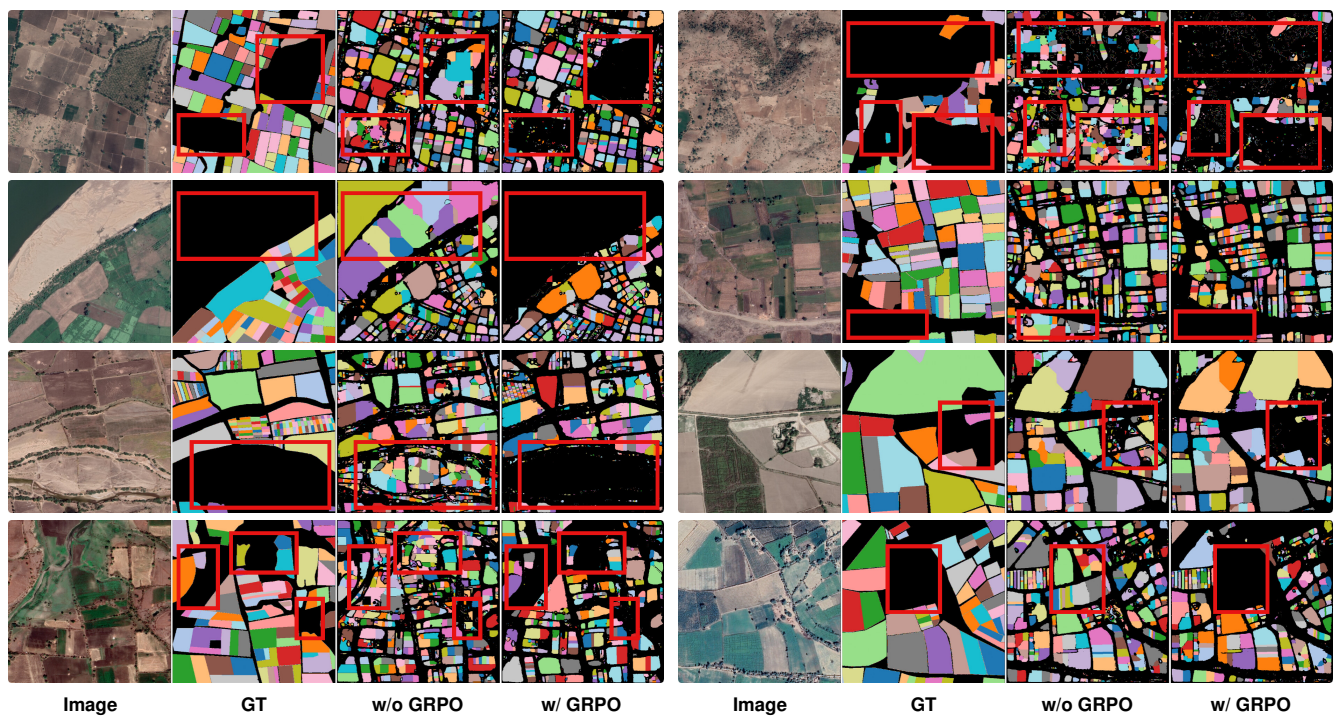}
    }
    \caption{Qualitative comparison of MAgSeg's segmentation performance with and without GRPO post-training. Region of interests are emphasized in red colored boxes.}
    \label{fig:grpo_qual_comp}
\end{figure*}

We also provide qualitative evidence for the effect and importance of Stage~2 (GRPO post-training) in Figure~\ref{fig:grpo_qual_comp}, where regions of interest are highlighted with red boxes. Visually, we observe that while standard SFT is able to detect generic edges, it often fails to distinguish between true field boundaries and other high-contrast structures (e.g., tree lines). This distinction is critical for downstream applications that rely on accurate field delineation. We attribute this limitation to the next-token prediction objective, which lacks explicit pixel-level semantic supervision and therefore does not directly encourage boundary correctness. In contrast, the GRPO stage optimizes a reward based on the pixel-level DICE score, explicitly incentivizing semantically aligned masks. The GRPO-trained model consequently produces cleaner, more selective field contours, suppressing spurious boundaries that the model, after SFT alone, tends to hallucinate.

\subsection{Qualitative Analysis: Additional Visualizations}
\label{sec:more_qual}

We provide additional qualitative results for samples from ALU and AI4SmallFarms datasets in Figures~\ref{fig:alu_qualitative_results},
\ref{fig:cambodia_qualitative_results},
\ref{fig:vietnam_qualitative_results}.

\begin{figure*}[t] % Use figure* for full-width in two-column papers
    \centering
    % Adjust the width if you want margins (e.g., 0.95\linewidth)
    \resizebox{\textwidth}{!}{ 
        \input{tikz/alu_qualitative_vis}
    }
    \caption{Qualitative results on India data. We compare our approach, MAgSeg against SOTA segmentation baselines. GT: Ground Truth.}
    \label{fig:alu_qualitative_results}
\end{figure*}

\begin{figure*}[t] % Use figure* for full-width in two-column papers
    \centering
    % Adjust the width if you want margins (e.g., 0.95\linewidth)
    \resizebox{\textwidth}{!}{ 
        \input{tikz/cambodia_qualitative_vis}
    }
    \caption{Qualitative results on Cambodia data. We compare our approach, MAgSeg against SOTA segmentation baselines. GT: Ground Truth.}
    \label{fig:cambodia_qualitative_results}
\end{figure*}

\begin{figure*}[t] % Use figure* for full-width in two-column papers
    \centering
    % Adjust the width if you want margins (e.g., 0.95\linewidth)
    \resizebox{\textwidth}{!}{ 
        \input{tikz/vietnam_qualitative_vis}
    }
    \caption{Qualitative results on Vietnam data. We compare our approach, MAgSeg, against SOTA segmentation baselines. GT: Ground Truth.}
    \label{fig:vietnam_qualitative_results}
\end{figure*}

\subsection{Qualitative Analysis: Limitations of MAgSeg}
\label{sec:vis_limitations}
\begin{figure*}[t] % Use figure* for full-width in two-column papers
    \centering
    % Adjust the width if you want margins (e.g., 0.95\linewidth)
    \resizebox{\textwidth}{!}{ 
        \input{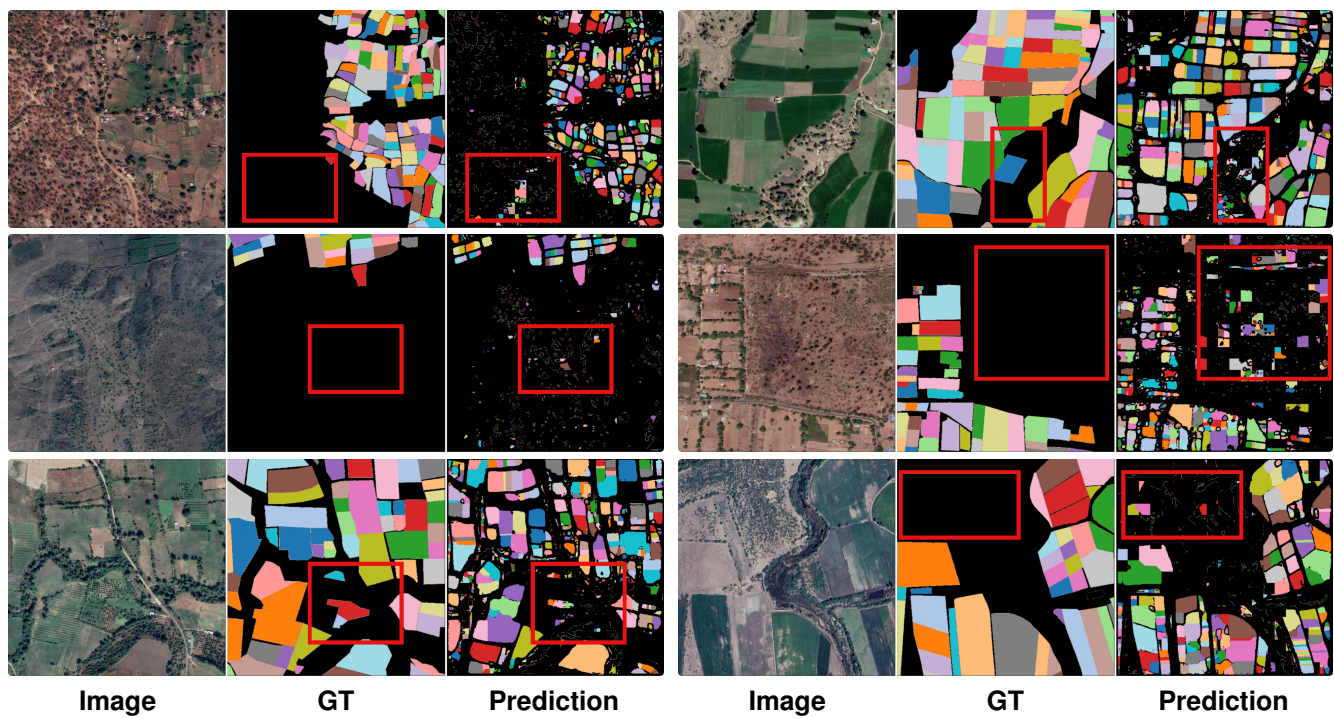}
    }
    \caption{Visualization of erroneous segmentation by MAgSeg. GT: Ground Truth. Region of interests are emphasized in red colored boxes.}
    \label{fig:error_analysis}
\end{figure*}

We provide qualitative examples of failure cases of the proposed method in Figure~\ref{fig:error_analysis}. In particular, we observe that for certain patches the model produces spurious masks that do not align with true field boundaries. We attribute this behaviour to the independent, patch-based inference strategy, where the model has no access to predictions from neighbouring patches. This lack of cross-patch context manifests as stitching artifacts and geometric discontinuities in the reconstructed segmentation maps.

\end{document}